\documentclass[lettersize,journal]{IEEEtran}
\usepackage{authblk}
\usepackage{amsmath,amsfonts}
\usepackage{algpseudocode}
\usepackage{algorithmicx}
\usepackage{amssymb}
\usepackage{array}
\usepackage{cuted}
\usepackage{mathtools}
\usepackage{hyperref}
\hypersetup{hidelinks}
\usepackage{pifont}
\usepackage{natbib}
\usepackage{balance}
\renewcommand{\triangleq}{\ensuremath{\mathrel{\overset{\Delta}{=}}}}
\usepackage{textcomp}
\usepackage{stfloats}
\usepackage[font=footnotesize,labelfont=bf]{caption}
\usepackage{subcaption}

\usepackage{url}
\usepackage{verbatim}
\usepackage{graphicx}
\usepackage{multirow}
\usepackage{placeins}

\newcolumntype{L}{>{\centering\arraybackslash}m{1.8cm}}

\usepackage{amsmath,amsfonts,bm}

\def\eqref#1{equation~\ref{#1}}

\def\1{\bm{1}}

\DeclareMathAlphabet{\mathsfit}{\encodingdefault}{\sfdefault}{m}{sl}
\SetMathAlphabet{\mathsfit}{bold}{\encodingdefault}{\sfdefault}{bx}{n}

\title{Hybrid Belief–Reinforcement Learning for Efficient Coordinated Spatial Exploration}
\author{Danish Rizvi}
\author{David Boyle}
{\vspace{-2mm}}
\affil{Imperial College London, UK}

{\vspace{-5mm}}

\begin{document}

\maketitle

\begin{abstract}
Coordinating multiple autonomous agents to explore and serve spatially heterogeneous demand requires jointly learning unknown spatial patterns and planning trajectories that maximize task performance. Pure model-based approaches provide structured uncertainty estimates but lack adaptive policy learning, while deep reinforcement learning often suffers from poor sample efficiency when spatial priors are absent. This paper presents a hybrid belief-reinforcement learning (HBRL) framework to address this gap. In the first phase, agents construct spatial beliefs using a Log-Gaussian Cox Process (LGCP) and execute information-driven trajectories guided by a Pathwise Mutual Information (PathMI) planner with multi-step lookahead. In the second phase, trajectory control is transferred to a Soft Actor-Critic (SAC) agent, warm-started through dual-channel knowledge transfer: i) belief state transfer supplies spatial uncertainty, and ii) replay buffer seeding provides demonstration trajectories generated during LGCP exploration. A variance-normalized overlap penalty enables coordinated coverage through shared belief state, permitting cooperative sensing in high-uncertainty regions while discouraging redundant coverage in well-explored areas. The framework is instantiated in two scenarios: multi-Unmanned Aerial Vehicle (UAV) wireless service provisioning, and mobility-on-demand sensing over measured New York City taxi demand. In the wireless scenario it achieves 10.8\% higher cumulative reward and takes 38\% fewer episodes to reach the reward level of the baselines, with ablation studies confirming that dual-channel transfer outperforms either channel alone; the taxi scenario yields consistent gains with all algorithmic components unchanged.
\end{abstract}

\section{Introduction}

Spatial exploration under uncertainty arises in numerous domains where autonomous agents must learn an unknown spatial field while simultaneously optimizing a task objective. Applications include environmental monitoring, dynamic wireless connectivity scaling, precision agriculture, disaster response, and infrastructure inspection~\citep{McEnroe2022SurveyEdgeAIUAV, Popovic2024LearningIPP}. A fundamental challenge is that the spatial field of interest is heterogeneous and \textit{a priori} unknown to the agents. Unlike settings where the spatial distribution is known or can be specified in advance, agents must simultaneously learn the underlying field while optimizing task performance. This gives rise to the coupled exploration-exploitation problem. 

Using multiple agents offers advantages including expanded coverage, increased redundancy, and load balancing. However, co-ordination between mobile agents introduces significant challenges, including: (i) joint trajectory optimization over high-dimensional continuous action spaces; (ii) coordinating multiple agents through shared spatial awareness; (iii) balancing information acquisition against task performance; and (iv) sample-efficient learning without prior knowledge of spatial structure. These challenges are compounded by partial observability, where agents can only identify demand within their immediate sensing region~\citep{Popovic2024LearningIPP}.

Two dominant paradigms have emerged. Model-based approaches employ spatial statistical models to maintain probabilistic beliefs, enabling principled uncertainty quantification~\citep{Krause2008Submodular}. However, these methods typically rely on myopic planning and do not adapt based on accumulated experience. Model-free deep reinforcement learning (DRL) can learn complex coordination policies, but suffers from poor sample efficiency~\citep{Wang2022DRLTrajectoryMEC}. This dichotomy motivates our work: \textit{combining the sample efficiency of Bayesian spatial modeling with the adaptive policy learning of deep RL}. Our framework combines two known constructs from the RL literature: the \emph{belief state}~\citep{kaelbling1998planning}, a summary of the agent's uncertainty over the environment, and \emph{learning from behavioural demonstrations}, which seeds policy learning with expert trajectories.

\subsection{Contributions}

We make the following contributions:

\begin{itemize}
    \item We propose a hybrid framework combining Log-Gaussian Cox Process (LGCP) spatial modeling with Soft Actor-Critic (SAC) reinforcement learning for coordinated spatial exploration under unknown demand.
    
    \item We introduce a dual-channel warm-start mechanism for sample-efficient learning: (i) \textit{belief state transfer}, which provides the RL agent with an informed prior for early policy updates, and (ii) \textit{behavioral transfer}, which seeds the replay buffer with LGCP-generate exploration trajectories. Ablation studies show that behavioral transfer provides the dominant benefit, while belief state transfer yields additional gains when combined with replay buffer seeding.
    
    \item We employ uncertainty-driven Pathwise Mutual Information (PathMI) planning for non-myopic trajectory optimization during the exploration phase, extending standard informative path planning (IPP) with staleness-weighted revisitation incentives.
    
    \item We propose a variance-normalized overlap penalty that adapts coordination strength to local belief uncertainty, permitting cooperative sensing in high-uncertainty regions, while penalizing redundant coverage.
    
    \item We instantiate and evaluate the framework in two structurally and functionally distinct scenarios: multi-UAV wireless service provisioning, achieving up to 10.8\% higher reward and 38\% fewer episodes to reach the reward level of the baselines, and mobility-on-demand sensing over New York City taxi demand (Appendix~\ref{Appendix_C}), where the unchanged framework yields consistent gains.\footnote{All code and research artifacts will be made openly available upon acceptance of the manuscript for publication. Additional details and updates can be found via the project page:~\href{https://smrizvi1.github.io/HBRL/}{https://smrizvi1.github.io/HBRL/}}
\end{itemize}

The remainder of the paper is organised as follows. Section~\ref{Related Work and Motivation} reviews related work and identifies five research gaps that motivate the proposed approach. Section~\ref{System Model} formalises the system model, including the spatial demand field, the LGCP belief model, and the reward structure. Section~\ref{Proposed Solution} presents the proposed HBRL framework, detailing PathMI planning, the dual-channel knowledge transfer mechanism, and the SAC training procedure. Section~\ref{sec:results} evaluates the framework through comprehensive ablation studies and comparisons with baselines. Section~6 concludes and discusses limitations and future directions.
\section{Related Work and Motivation}
\label{Related Work and Motivation}

\subsection{Informative Path Planning}

Informative path planning (IPP) addresses trajectory optimization for information acquisition under uncertainty. \cite{Krause2007NonmyopicGP} propose an exploration–exploitation framework for single-agent sequential observation selection under unknown Gaussian Process (GP) kernel parameters. Relatedly, \cite{Krause2008Submodular} prove that mutual information for GP regression is submodular, yielding a greedy approximate algorithm for static multi-sensor placement. Both, however, are restricted to discrete candidate sets and do not extend to continuous action spaces for mobile robot path planning.                                                  
                                                               
\cite{cao2023catnipp} propose CAtNIPP, a learning-based framework for adaptive informative path planning that uses attention mechanisms to encode a global belief and guide non-myopic decisions, but in a single-agent setting only. \cite{Chen2024AdaptiveNonstationaryGP} address non-stationary GPs for robotic information gathering, again considering single-agent scenarios only. \cite{stephens2024planning} incorporate safety constraints into GP-based exploration, but do not address coordination among multiple agents. While these methods provide principled uncertainty-driven exploration, they predominantly focus on single-agent scenarios with pre-specified GP
hyperparameters.

\subsection{Spatial Statistical Modeling}

For count data exhibiting spatial correlation, Log-Gaussian Cox Processes (LGCPs) model spatially varying intensity fields with Bayesian uncertainty quantification~\citep{Moller1998LogGaussianCox}. Computational tractability is achieved through Gaussian Markov Random Field (GMRF) approximations~\citep{Rue2005GMRF, Lindgren2011SPDE} and Integrated Nested Laplace Approximation (INLA)~\citep{Rue2009INLA}. \cite{Diggle2013SpatialLGCP} provides comprehensive treatment of LGCPs for spatial point patterns.

Recent advances include proposing fast maximum likelihood estimation using automatic differentiation~\citep{Dovers2024FastMLELGCP}, and demonstrating LGCP fitting via GAM software~\citep{dovers2024fitting}. LGCPs are widely used for spatial point process modeling, particularly in environmental and epidemiological applications~\citep{JonesTodd2024Stelfi,DAngelo2022LocalLGCP}. Although, LGCPs provide principled uncertainty quantification for count data, their integration with robotic path planning or reinforcement learning remains relatively underexplored.

\subsection{Reinforcement Learning-Based Coordination}

Reinforcement learning (RL) has emerged as a promising paradigm for coordinated spatial exploration, with UAV coordination being a prominent application. The survey by~\cite{Bai2023SurveyMARLUAV} identifies sample efficiency as a critical challenge. \cite{Wang2022DRLTrajectoryMEC} apply DRL for trajectory design in mobile edge computing (MEC) systems, demonstrating improved performance but reporting slow convergence.

Heterogeneous multi-agent reinforcement learning (MARL) has recently been proposed for joint trajectory and communication design but does not incorporate spatial priors~\citep{Zhou2024HeteroMARL}. \cite{rizvi2024multi} develop shared Deep Q-Network (DQN) with action masking for Non-Orthogonal Multiple Access (NOMA)-UAV networks noting that fixing cluster sizes limits adaptability to varying user distributions. \cite{Liu2024GraphMARLSearchTracking} employ graph-based MARL for search and tracking but focus on target localization rather than demand learning. Without structured spatial priors, these RL-based coordination methods suffer from poor sample efficiency in unknown environments. Scaling joint-action formulations to substantially larger fleets typically
requires parameter sharing, decentralized execution, or graph-based policy
architectures \citep{Liu2024GraphMARLSearchTracking}; these address a different design dimension
from the belief-guided warm start proposed here and are complementary to it.

\subsection{Transfer Learning and Warm-Starting}

Transfer learning accelerates RL by leveraging knowledge from related tasks or demonstrations. \cite{Hester2018DQfD} propose Deep Q-learning from Demonstrations, seeding replay buffers with expert trajectories, and \cite{ball2023efficient} retain a pre-collected offline dataset in a separate buffer sampled alongside online data. However, this requires access to prior demonstrations unavailable in novel environments. The survey by~\cite{Zhu2023TransferRL} identifies domain adaptation as a key challenge.

\cite{Skrynnik2021ForgetfulER} address forgetful experience replay for hierarchical RL but focus on single-agent navigation. \cite{Wang2024FlexNetWarmStart} demonstrate warm-starting for hybrid vehicle energy management, showing significant sample efficiency gains. While these methods improve sample efficiency, the use of structured Bayesian spatial priors as the knowledge source for coordinated spatial exploration remains unexplored. Unlike conventional demonstration-based warm starts that rely on externally provided expert trajectories, HBRL generates informative trajectories during its belief-guided PathMI warm-start and transfers both the resulting behavioural experience and spatial belief information to SAC.

\subsection{Trajectory Optimization}

Trajectory design for mobile agents, particularly in UAV-enabled communications, has received substantial attention. \cite{Zeng2019EnergyMinimization} minimize propulsion energy for rotary-wing UAVs using successive convex approximation but assume perfect knowledge of user positions throughout the mission. \cite{Wu2018JointTrajectoryComm} jointly optimizes trajectory and power allocation; however, the approach requires full channel state information which may not always be available. \cite{Hao2024TaskOffloadingPriority} address task offloading with trajectory control but do not incorporate online demand learning. \cite{Song2023EMORLMEC} apply evolutionary multi-objective RL for UAV-assisted MEC but focus on computational offloading rather than demand learning. These methods assume known demand distributions, or employ full state knowledge, rendering them unsuitable when demand must be learned online from partial observations.

\subsection{Research Gaps and Motivation}

The reviewed literature reveals five key limitations that motivate our work:

\begin{itemize}
    \item Classical trajectory optimization assumes known user distributions~\citep{Zeng2019EnergyMinimization, Wu2018JointTrajectoryComm, Hao2024TaskOffloadingPriority}. In many practical deployments, demand patterns are initially unknown and must be learned online.

    \item Many IPP methods employ greedy one-step planning~\citep{Krause2007NonmyopicGP}. While computationally efficient, myopic strategies fail to anticipate future information gain, leading to suboptimal long-horizon performance.

    \item DRL approaches lack principled uncertainty estimates~\citep{Wang2022DRLTrajectoryMEC, Zhou2024HeteroMARL, rizvi2024multi}. Without calibrated uncertainty, agents cannot distinguish ``unexplored'' from ``low-demand'' regions.

    \item Pure DRL suffers from poor sample efficiency without spatial priors~\citep{Bai2023SurveyMARLUAV}. Agents must discover demand structure through extensive trial and error.

    \item Much of the IPP and spatial modeling literature focuses on single-agent scenarios~\citep{Popovic2024LearningIPP, Chen2024AdaptiveNonstationaryGP}. Coordination among multiple agents requires redundant coverage avoidance and cooperative sensing strategies, which are not addressed by these methods.

\end{itemize}

\begin{table}[!t]
\caption{Summary of related work and key limitations. 
UD: Unknown Demand, MA: Multiple mobile Agents, NM: Non-Myopic Planning, 
SU: Spatial Uncertainty, SE: Sample Efficiency.}
\label{tab:related_work}
\renewcommand{\arraystretch}{1.2}
\setlength{\tabcolsep}{5pt}
\centering
\footnotesize
\begin{tabular}{p{2cm} p{2.4cm} c c c c c}
\hline
\textbf{Reference} & \textbf{Approach} & \textbf{UD} & \textbf{MA} & \textbf{NM} & \textbf{SU} & \textbf{SE} \\
\hline

~\cite{Krause2008Submodular}
& Greedy Submodular placement
& \ding{51} & \ding{55} & \ding{55} & \ding{51} & n/a \\

~\cite{Wu2018JointTrajectoryComm}
& Joint traj.\ \& resource
& \ding{55} & \ding{51} & \ding{51} & \ding{55} & n/a \\

~\cite{Zeng2019EnergyMinimization}
& Convex Opt.\ Min.\ flight energy
& \ding{55} & \ding{55} & \ding{51} & \ding{55} & n/a \\

~\cite{Wang2022DRLTrajectoryMEC}
& DRL traj.\ for MEC
& \ding{51} & \ding{51} & \ding{51} & \ding{55} & \ding{55} \\

~\cite{cao2023catnipp}
& Learning-based IPP
& \ding{51} & \ding{55} & \ding{51} & \ding{51} & \ding{51} \\

~\cite{Zhou2024HeteroMARL}
& Heterogeneous MARL
& \ding{51} & \ding{51} & \ding{51} & \ding{55} & \ding{55} \\

~\cite{Hao2024TaskOffloadingPriority}
& Task offloading \& traj.
& \ding{55} & \ding{51} & \ding{51} & \ding{55} & \ding{55} \\

~\cite{rizvi2024multi}
& SDQN + action masking
& \ding{55} & \ding{51} & \ding{51} & \ding{55} & \ding{55} \\

\textbf{This Work}
& HBRL (LGCP--SAC)
& \ding{51} & \ding{51} & \ding{51} & \ding{51} & \ding{51} \\

\hline
\end{tabular}
\end{table}

As shown in Table~\ref{tab:related_work}, prior works fail to address all requirements simultaneously. Our HBRL framework bridges this gap through: (i) LGCP-based Bayesian beliefs over unknown demand, (ii) PathMI non-myopic planning, (iii) dual-channel knowledge transfer for sample efficiency, and (iv) variance-normalized penalties for coordinated coverage.

\section{System Model}
\label{System Model}

\subsection{System Description}

\begin{figure*}[!t]
    \centering
    \includegraphics[viewport=020 020 380 193, clip, width=4in]{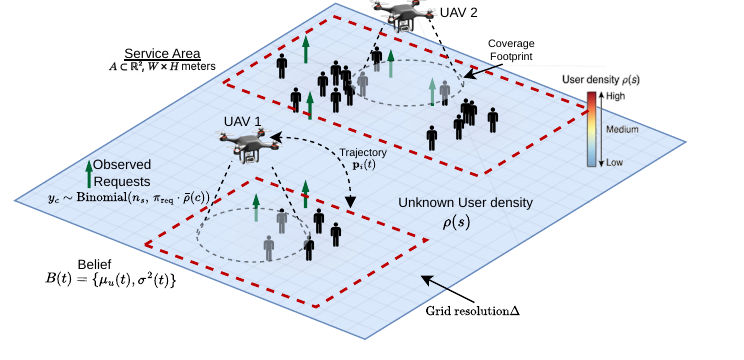}
    \caption{Multiple mobile agents spatial exploration over a discretized operational area with unknown demand, modeled via a spatial belief process. The illustrated scenario depicts UAVs as the instantiated agents, with further details provided in Section~4.}
    \label{system_model}
\end{figure*}
As shown in Figure~\ref{system_model}, we consider a spatial exploration task where $N$ autonomous agents are deployed to serve demand distributed across a two-dimensional service area $\mathcal{A} \subset \mathbb{R}^2$ of dimensions $L_x \times L_y$ meters. To enable tractable spatial modeling, the area is discretized into a uniform grid with spatial resolution $\Delta$ meters, yielding $G_x \times G_y$ cells, where $G_x = \lfloor L_x/\Delta \rfloor$ and $G_y = \lfloor L_y/\Delta \rfloor$.

Each agent $i \in \{1,\ldots,N\}$ operates within a predefined operational subregion $\mathcal{A}_i \subset \mathcal{A}$. Subregions may overlap. Each agent senses demand within a circular footprint of radius $r_c$ meters, centered at position $\mathbf{p}_i(t) = (x_i(t), y_i(t))^\top$ at discrete time step $t$. All quantities indexed by $c \in \mathcal{G}$ are computed over the operational grid $\mathcal{G} = \bigcup_{i=1}^{N} \mathcal{G}_i$, the union of discretized subregions, rather than over the full service area.

User demand is assumed to be spatially heterogeneous and concentrated around $J$ hotspot regions whose locations and intensities are initially unknown to the agents. This assumption reflects realistic scenarios where event clusters around spatial hotspots whose structure must be discovered online. The mission objective is to jointly optimize agent trajectories over a horizon of $T$ time steps to maximize cumulative task reward while efficiently learning the underlying spatial demand distribution.

\begin{table*}[!t]
\caption{Summary of Key Notations}
\label{tab:notations}
\begin{center}
\begin{tabular}{ll@{\hspace{0.6cm}}ll@{\hspace{0.6cm}}ll}
\textbf{Symbol} & \textbf{Description} &
\textbf{Symbol} & \textbf{Description} &
\textbf{Symbol} & \textbf{Description} \\
\hline \\[-0.9em]

$N$ & Number of agents &
$T$ & Mission horizon &
$\mathcal{A}$ & Service area \\

$\mathcal{G}$ & Discretized grid &
$\Delta$ & Grid resolution &
$G_x, G_y$ & Grid dimensions \\

$\mathbf{p}_i(t)$ & Position of agent $i$ &
$d_{\max}$ & Max.\ displacement &
$r_c$ & Sensing radius \\

$\mathcal{C}(t)$ & Covered cells at $t$ &
$y_c(t)$ & Observed event count &
$N_{\text{req}}(t)$ & Total requests \\

$\rho(\mathbf{s})$ & Ground-truth density &
$\bar{\rho}(c)$ & Cell-averaged density &
$\boldsymbol{\mu}_j$ & Hotspot $j$ center \\

$\sigma_j$ & Hotspot $j$ spread &
$\sigma_h$ & Hotspot diffusion rate &
$r_b$ & Displacement bound \\

$\mathbf{u}$ & Log-intensity field &
$\boldsymbol{\lambda}$ & Intensity field &
$\sigma_c^2(t)$ & Posterior variance \\

$\tilde{\sigma}_c^2(t)$ & Predicted variance &
$e_c(t)$ & Exposure indicator &
$n_c$ & Visit count \\

$s_c$ & Staleness counter &
$s_{\max}$ & Max.\ staleness &
$\tau$ & Prior precision \\

$\beta$ & Spatial smoothness &
$\gamma_v$ & Variance growth rate &
$\sigma^2_{\max}$ & Max.\ variance \\

$r_g$ & Coverage radius in cells &
$\sigma^2_{\min}$ & Min.\ variance &
$L$ & Planning horizon \\

$D_{dir}$ & Candidate directions &
$\xi_c$ & Diminishing return &
$\epsilon$ & Stability constant \\

$m_c(t)$ & Overlap count &
$K_w$ & Warm-start episodes &
$\omega_1,\omega_2,\omega_3$ & Reward weights \\

$\omega_s$ & Staleness weight &
$\gamma$ & Discount factor &
$\tau_{\text{soft}}$ & Soft update coeff. \\

\end{tabular}
\end{center}
\end{table*}
This setting requires balancing exploration of unknown regions against exploitation of discovered high-demand areas, a fundamental challenge in sequential decision-making under uncertainty~\citep{Krause2007NonmyopicGP}. A summary of the key notations used throughout the paper is provided in Table~\ref{tab:notations}.

\subsection{Agent Dynamics Model}

Each agent is modeled as a point mass entity with discrete-time dynamics. At each time step $t$, agent $i$ selects a displacement vector $\mathbf{v}_i(t) \in \mathbb{R}^2$. The position update is given by
\begin{equation}
    \mathbf{p}_i(t+1) = \text{clip}\!\left(\mathbf{p}_i(t) + \mathbf{v}_i(t), \, \mathcal{A}\right),
    \label{eq:position_update}
\end{equation}
where $\text{clip}(\cdot, \mathcal{A})$ enforces boundary constraints such that $0 \leq x_i(t) \leq L_x$ and $0 \leq y_i(t) \leq L_y$.

The control input for agent $i$ at time $t$ is specified as a continuous action $\mathbf{a}_i(t) \in [-1,1]^2$, mapped to displacement via
\begin{equation}
    \mathbf{v}_i(t) = \mathbf{a}_i(t) \cdot d_{\max},
    \label{eq:action_mapping}
\end{equation}
where $d_{\max}$ denotes the maximum displacement per time step. Since $\mathbf{a}_i(t) \in [-1,1]^2$, this mapping enforces an $\ell_\infty$ bound on displacement, i.e. $\|\mathbf{v}_i(t)\|_\infty \leq d_{\max}$. For a fleet of $N$ agents, the joint action is $\mathbf{a}(t) = [\mathbf{a}_1(t)^\top, \ldots, \mathbf{a}_N(t)^\top]^\top \in [-1,1]^{2N}$.

The sensing footprint of agent $i$ at time $t$ is defined as the set of grid cells within sensing range:
\begin{equation}
    \mathcal{C}_i(t) = \left\{ c \in \mathcal{G} : \|\mathbf{p}_c - \mathbf{p}_i(t)\|_2 \leq r_c \right\},
    \label{eq:coverage_footprint}
\end{equation}
where $\mathbf{p}_c \in \mathbb{R}^2$ denotes the center position of grid cell $c$. The joint coverage at time $t$ is $\mathcal{C}(t) = \bigcup_{i=1}^{N} \mathcal{C}_i(t)$.

\subsection{Spatial Demand Model}

 Demand is modeled as a spatially heterogeneous density field over the operational area. The ground-truth normalized density $\rho : \mathcal{A} \rightarrow [0,1]$ is generated as a mixture of $J$ Gaussian hotspots:
\begin{equation}
    \rho(\mathbf{s}) = \frac{1}{\rho_{\max}} \sum_{j=1}^{J}
    \exp\!\left( -\frac{\|\mathbf{s} - \boldsymbol{\mu}_j\|_2^2}{2\sigma_j^2} \right),
    \label{eq:density_field}
\end{equation}
where $\boldsymbol{\mu}_j \in \mathcal{A}$ and $\sigma_j > 0$ denote the center and spatial spread of hotspot $j$, and $\rho_{\max}$ is a normalization constant ensuring $\rho(\mathbf{s}) \in [0,1]$. To model realistic local demand fluctuations, hotspot centers undergo 
a bounded random walk during each episode:
\begin{equation}
    \boldsymbol{\mu}_j(t+1) = \text{clip}\left(
        \boldsymbol{\mu}_j(t) + \sigma_h \boldsymbol{\epsilon}_t, \;
        \boldsymbol{\mu}_j^{(0)} \pm r_b
    \right),
    \label{eq:hotspot_drift}
\end{equation}
where $\sigma_h > 0$ is the diffusion rate, $\boldsymbol{\epsilon}_t \sim \mathcal{N}(\mathbf{0}, \mathbf{I})$ denotes isotropic Gaussian noise, $\boldsymbol{\mu}_j^{(0)}$ is the base hotspot position, and $r_b$ bounds the maximum displacement. At episode reset, hotspot centers return to their base positions with small perturbation, preserving spatial structure across episodes while introducing within-episode non-stationarity.

The mobile agents have no prior knowledge of the hotspot parameters $\{\boldsymbol{\mu}_j, \sigma_j\}_{j=1}^{J}$ or its position, necessitating online learning of the spatial demand pattern from partial observations.

\textit{Remark 1 (Density vs.\ Belief):} The density $\rho(\mathbf{s})$ represents the ground-truth spatial distribution used to generate observations. The LGCP intensity $\lambda(\mathbf{s}) = \exp(u(\mathbf{s}))$ is the agents' learned belief over event rates. While these quantities differ in scale and interpretation, regions with higher $\rho(\mathbf{s})$ induce higher learned intensity in the belief model.

\subsection{Observation Model}

Agents can observe events only within their current sensing regions. At each time step $t$, for each grid cell $c \in \mathcal{C}(t)$, the observed event count is sampled as
\begin{equation}
    y_c(t) \sim \text{Binomial}\!\left(n_s, \, \pi_{\text{req}} \, \bar{\rho}(c)\right),
    \label{eq:observation_model}
\end{equation}
where $n_s$ denotes the number of sampling trials, $\pi_{\text{req}} \in (0,1]$ is the event, or request, probability, and $\bar{\rho}(c)$ is the average normalized density over cell $c$, computed as
\begin{equation}
    \bar{\rho}(c) = \frac{1}{|\mathcal{A}_c|} \int_{\mathcal{A}_c} \rho(\mathbf{s}) \, d\mathbf{s},
    \label{eq:cell_density}
\end{equation}
where $\mathcal{A}_c \subset \mathcal{A}$ denotes the spatial region corresponding to cell $c$ and $|\mathcal{A}_c| = \Delta^2$ is the cell area. Cells not covered at time $t$ yield no observations.

The total observed count at time $t$ is
\begin{equation}
    N_{\text{req}}(t) = \sum_{c \in \mathcal{C}(t)} y_c(t).
    \label{eq:total_requests}
\end{equation}

\textit{Remark 2 (Likelihood Approximation):} While observations are generated from a binomial model, a Poisson likelihood is employed during belief inference. With small success probability ($\pi_{\text{req}} \leq 0.05$), this approximation is well justified under classical Poisson limit results~\citep{LeCam1960Poisson}.

\subsection{Log-Gaussian Cox Process Belief Model}

We adopt a Log-Gaussian Cox Process (LGCP) for spatial demand estimation, following the formulation of M{\o}ller et al.~\citep{Moller1998LogGaussianCox} with grid-based Laplace-approximated inference~\citep{Rue2009INLA}. Our contribution lies in its online application under partial observability with temporal belief decay, and its integration with information-driven planning and reinforcement learning for coordinating multiple mobile agents.

\subsubsection{LGCP Formulation}
Following standard LGCP notation~\citep{Diggle2013SpatialLGCP}, the log-intensity is modeled as a Gaussian process:
\begin{equation}
    \lambda(\mathbf{s}) = \exp(u(\mathbf{s})), \quad u(\mathbf{s}) \sim \mathcal{GP}(0, \mathcal{K}),
    \label{eq:lgcp_definition}
\end{equation}
where $u : \mathcal{A} \rightarrow \mathbb{R}$ is the latent log-intensity field and $\mathcal{K}$ denotes a covariance kernel encoding spatial correlation. The exponential link ensures positive intensity values, while the Gaussian prior induces a spatially correlated uncertainty structure~\citep{Moller1998LogGaussianCox}.

\subsubsection{GMRF Prior}
For computational tractability on the discretized grid $\mathcal{G}$, a Gaussian Markov Random Field (GMRF) prior is imposed~\citep{Rue2005GMRF}, providing sparse precision matrices for efficient inference. Let $\mathbf{u} \in \mathbb{R}^{G_x \cdot G_y}$ denote the vectorized log-intensity field over all grid cells.

Following standard GMRF constructions~\citep{Rue2005GMRF}, the prior is specified through its precision matrix:
\begin{equation}
    \mathbf{Q} = \tau \mathbf{I} + \beta \mathbf{L}_G,
    \label{eq:precision_matrix}
\end{equation}
where $\tau > 0$ controls prior precision (regularization strength), $\beta > 0$ governs spatial smoothness, $\mathbf{I}$ is the identity matrix, and $\mathbf{L}_G$ denotes the graph Laplacian encoding four-connected neighbor relationships on the grid. For a grid cell with four neighbors, the graph Laplacian has diagonal entries $[\mathbf{L}_G]_{cc} = 4$ (the node degree) and off-diagonal entries $[\mathbf{L}_G]_{cj} = -1$ for neighboring cells $j$. This formulation induces the prior distribution $p(\mathbf{u}) \propto \exp\left(-\frac{1}{2}\mathbf{u}^\top \mathbf{Q} \mathbf{u}\right)$, which is proper and positive definite for $\tau > 0$.

\subsubsection{Posterior Inference via Laplace Approximation}
\label{sec:laplace}

Given observations $\mathbf{y}(t) = \{y_c(t)\}_{c:\, e_c(t)=1}$ (Eq.~\ref{eq:exposure_indicator}) from previously covered cells, where $y_c(t)$ is the count from cell $c$'s most recent observation, the posterior distribution over $\mathbf{u}$ combines the GMRF prior with the Poisson likelihood. The resulting posterior is analytically intractable due to the non-conjugate Poisson likelihood. We employ a Laplace approximation via Newton iterations~\citep{Dovers2024FastMLELGCP}, yielding a Gaussian approximation centered at the posterior mode:
\begin{equation}
    p(\mathbf{u}|\mathbf{y}(t)) \approx \mathcal{N}(\boldsymbol{\mu}_u, \boldsymbol{\Sigma}_u),
    \label{eq:laplace_approx}
\end{equation}
where $\boldsymbol{\mu}_u$ denotes the posterior mode and $\boldsymbol{\Sigma}_u = \mathbf{H}^{-1}$ is the posterior covariance, with $\mathbf{H}$ being the Hessian of the negative log-posterior.

To handle partial observability, we introduce a binary exposure variable $e_c(t) \in \{0, 1\}$ indicating whether cell $c$ is observed by time $t$ :
\begin{equation}
  e_c(t) = {\mathbb{I}\big(\exists\, t' \le t : c \in \mathcal{C}(t')\big)}
  \label{eq:exposure_indicator}
\end{equation}
where $\mathbb{I}(\cdot)$ is the indicator function. For cells not covered yet, $y_c(t) = 0$ by convention, with the exposure mask nullifying their likelihood contribution. The log-posterior at time $t$ then becomes:
\begin{equation}
    \log p(\mathbf{u}|\mathbf{y}(t)) \propto \sum_{c \in \mathcal{G}} e_c(t) \left[y_c(t) u_c - \exp(u_c)\right] - \frac{1}{2}\mathbf{u}^\top \mathbf{Q} \mathbf{u},
    \label{eq:log_posterior}
\end{equation}
where $e_c(t) = 1$ for previously covered cells and $e_c(t) = 0$ otherwise. The gradient and Hessian follow as:
\begin{align}
    \mathbf{g}(t) &= \mathbf{e}(t) \odot (\mathbf{y}(t) - \boldsymbol{\lambda}) - \mathbf{Q}\mathbf{u}, \label{eq:gradient} \\
    \mathbf{H}(t) &= \mathrm{diag}(\mathbf{e}(t) \odot \boldsymbol{\lambda}) + \mathbf{Q}, \label{eq:hessian}
\end{align}
where $\boldsymbol{\lambda} = \exp(\mathbf{u})$ is the intensity vector, $\mathbf{e}(t)$ denotes the exposure vector at time $t$, and $\odot$ represents element-wise multiplication. The Newton update is:
\begin{equation}
    \mathbf{u}^{(k+1)} = \mathbf{u}^{(k)} + \mathbf{H}(t)^{-1} \mathbf{g}(t).
    \label{eq:newton_update}
\end{equation}
The Newton iterations are warm-started from the previous Maximum A Posteriori (MAP) estimate, $\mathbf{u}^{(0)} = \boldsymbol{\mu}_u(t-1)$ . 

A preconditioned conjugate gradient (PCG) solver~\citep{Rue2005GMRF} computes the Newton step $\mathbf{H}(t)^{-1}\mathbf{g}(t)$ without explicit matrix inversion. The Hessian admits diagonal entries:
\begin{equation}
    [\mathbf{H}(t)]_{cc} = e_c(t)\lambda_c + Q_{cc},
    \label{eq:hessian_diagonal}
\end{equation}
where $Q_{cc} = \tau + 4\beta$ for interior cells under the first-order GMRF structure. We use these diagonal entries as a preconditioner $\mathbf{M} = \mathrm{diag}(\mathbf{H})$, which exploits the sparsity induced by the GMRF prior and accelerates PCG convergence.

Computing exact marginal posterior variances $\mathrm{diag}(\mathbf{H}^{-1})$ requires sparse matrix inversion, which is prohibitive for large grids~\citep{Rue2009INLA}. Instead, we employ a diagonal Laplace approximation~\citep{ritter2018scalable,kirkpatrick2017overcoming}:

\begin{equation}
    \sigma_c^2(t)\triangleq\frac{1}{[\mathbf{H}(t)]_{cc}} .
    \label{eq:diagonal_variance}
\end{equation}

This surrogate sacrifices calibrated marginal variances but preserves relative uncertainty ordering across cells, which makes it suitable for exploration incentives where ranking drives decision-making. For brevity, we refer to this quantity as the \textit{posterior variance} in the remainder of the paper, noting that it corresponds to a diagonal Laplace approximation rather than the exact marginal variance. A direct comparison against the full marginal variance is tractable at small grid scales, but the cost of exact inversion renders such comparison impractical at the operational grid size in real world scenarios. The proxy is used because it preserves the relative uncertainty ordering on which exploration decisions depend.

Appendix~\ref{Appendix_D} quantifies this agreement empirically on grids small enough for exact inversion, confirming that both cell-level ordering and the resulting PathMI action selection are preserved.

For observed cells ($e_c(t) = 1$), higher intensity estimates yield smaller variance proxies, reflecting increased confidence. For unobserved cells ($e_c(t) = 0$), the likelihood contributes no curvature, and uncertainty evolution is governed solely by the temporal belief dynamics described next.

\subsection{Temporal Belief Dynamics}
\label{sec:temporal}
In partially observable environments, the relevance of past observations degrades over time due to limited and intermittent sensing. We introduce a two-step belief update that separates uncertainty growth from observation-based refinement.

\subsubsection{Step 1: Prediction (Uncertainty Growth)}
At each time step, before incorporating new observations, we apply variance growth to all cells to reflect potential changes in the underlying demand pattern:
\begin{equation}
    \tilde{\sigma}_c^2(t) = \min\!\left(\sigma_c^2(t^{-}) \cdot (1 + \gamma_v), \, \sigma^2_{\max}\right), \quad \forall c \in \mathcal{G},
    \label{eq:variance_predict}
\end{equation}
where $\sigma_c^2(t^{-})$ denotes the posterior variance at the end of the previous time step $t-1$, $\gamma_v > 0$ is the variance growth rate per time step, and $\sigma^2_{\max}$ represents the maximum variance corresponding to complete uncertainty.

\subsubsection{Step 2: Update (Observation Incorporation)}
After the prediction step, we incorporate observations from currently covered cells. For observed cells, the variance is overwritten by the posterior variance from LGCP inference; for unobserved cells, the predicted variance is retained:
\begin{equation}
    \sigma_c^2(t^{+}) = 
    \begin{cases}
        \max\!\left(\dfrac{1}{\lambda_c^* + \tau + 4\beta}, \, \sigma^2_{\min}\right) & \text{if } c \in \mathcal{C}(t) \\[10pt]
        \tilde{\sigma}_c^2(t) & \text{otherwise}
    \end{cases},
    \label{eq:variance_update}
\end{equation}
where $\lambda_c^* = \exp(u_c^*)$ is the converged posterior mode intensity and $\sigma^2_{\min} > 0$ is a small floor preventing numerical instability in high-intensity regions.

This predict-then-update structure serves two purposes: (i) it acknowledges that demand patterns may evolve, making older observations 
less reliable, and (ii) it incentivizes revisitation of stale regions where belief confidence has degraded. The belief state \citep{kaelbling1998planning} at time $t$ is
thus characterized by $\mathcal{B}(t) = \{\boldsymbol{\mu}_u(t), \boldsymbol{\sigma}^2(t^{+})\}$.

\textit{Remark 3 (Revisitation Incentive):} Under this formulation, 
revisiting a cell $c$ at time $t$ yields fresh observations $y_c(t)$ and resets the posterior variance to the observation-based value. This creates a natural incentive for revisitation: cells not recently observed accumulate uncertainty via Equation~\ref{eq:variance_predict}, making them attractive targets for exploration reward. Cells recently observed maintain low variance until staleness accumulates.

\subsection{Reward Structure}

The reward at time step \textit{t} jointly maximizes service performance and information acquisition through exploration, while minimizing coordination cost, and is calculated as:
\begin{equation}
    r_t = \omega_1 R_{\text{service}}(t) + \omega_2 R_{\text{explore}}(t) - \omega_3 C_{\text{coord}}(t),
    \label{eq:reward}
\end{equation}
where $\omega_1, \omega_2, \omega_3 > 0$ are objective weights. 

\subsubsection{Service Reward}
The service reward captures the primary mission objective:
\begin{equation}
    R_{\text{service}}(t) = N_{\text{req}}(t),
    \label{eq:service_reward}
\end{equation}
where $N_{\text{req}}(t)$ denotes the total observed event count at time $t$ as defined in Equation~\ref{eq:total_requests}. This term incentivizes agents to position themselves over high-demand regions.

\subsubsection{Exploration Reward}
The exploration reward encourages belief improvement through uncertainty reduction:
\begin{equation}
    R_{\text{explore}}(t) = \sum_{c \in \mathcal{C}(t)} \left(\tilde{\sigma}_c^2(t) - \sigma_c^2(t^{+})\right),
    \label{eq:explore_reward}
\end{equation}
where $\tilde{\sigma}_c^2(t)$ denotes the predicted variance (before observation) and $\sigma_c^2(t^{+})$ denotes the updated posterior variance (after observation) at cell $c$.

This formulation creates an implicit incentive mechanism: cells with high predicted variance (either unvisited or stale) yield large uncertainty reduction upon observation; recently visited cells with low variance provide minimal gain. Consequently, agents are naturally drawn toward unexplored or stale regions without requiring added revisitation penalties.

\subsubsection{Coordination Cost}
The coordination cost penalizes inefficient trajectories through redundant coverage and excessive travel:
\begin{equation}
    C_{\text{coord}}(t) = P_{\text{overlap}}(t) + d_{\text{travel}}(t),
    \label{eq:coord_cost}
\end{equation}
where
\begin{equation}
    d_{\text{travel}}(t) = \delta_t \sum_{i=1}^{N} \|\mathbf{p}_i(t) - \mathbf{p}_i(t-1)\|_2,
    \label{eq:travel_cost}
\end{equation}
with $\delta_t > 0$ serving as a scaling constant for dimensional consistency.

To avoid overlapping coverage, we use a variance-normalized overlap penalty. For each grid cell $c \in \mathcal{G}$, let $m_c(t) = |\{i : c \in \mathcal{C}_i(t)\}|$ denote the number of agent coverage footprints that include cell $c$ at time $t$. The overlap penalty is then defined as
\begin{equation}
    P_{\text{overlap}}(t) = \sum_{c: m_c(t) \geq 2} \left(1 - \frac{\tilde{\sigma}_c^2(t)}{\sigma^2_{\max}}\right),
    \label{eq:overlap_penalty}
\end{equation}
where $\tilde{\sigma}_c^2(t)$ is the predicted variance (before incorporating current observations), and $\sigma^2_{\max}$ represents complete uncertainty.

\textit{Remark 4 (Penalty Interpretation):} The use of predicted variance $\tilde{\sigma}_c^2(t)$ allows the penalty to adapt to belief state. In recently observed regions ($\tilde{\sigma}^2_c(t) \approx 1/(\lambda^*_c + \tau + 4\beta)$), the penalty is high, discouraging redundant coverage, whereas in unexplored or stale regions ($\tilde{\sigma}_c^2(t) \approx \sigma^2_{\max}$) it is near zero, permitting collaborative sensing. Overlap is hence penalized more heavily on recently covered cells than on cells not yet covered. This adaptive behavior allows joint exploration when uncertainty is high, but persistent overlap is discouraged once variance is reduced.

\begin{table}[!t]
\caption{Comparison of overlap penalty strategies for a cell covered by two agents.}
\label{tab:overlap_comparison}
\begin{center}
\begin{tabular}{lccc}
\hline \\[-0.9em]
\textbf{Cell History} & $\tilde{\sigma}_c^2(t)$ & \textbf{Var.-Norm.} & \textbf{Fixed} \\
\hline \\[-0.9em]

Never visited & 1.0 & 0.0 & 1.0 \\
Recently visited & 0.1 & 0.9 & 1.0 \\
Stale (grown variance) & 0.5 & 0.5 & 1.0 \\

\end{tabular}
\end{center}
\end{table}
A fixed overlap penalty applies uniform cost regardless of belief state and discourages all overlap equally, including beneficial joint exploration in high-uncertainty regions. Table~\ref{tab:overlap_comparison} summarizes this difference.


The overlap penalty is triggered only when $m_c(t) \geq 2$. Single-agent revisitation penalty is governed by the exploration reward and service objective, allowing agents to revisit high-demand regions when beneficial.

\subsubsection{Weight Selection}
The default objective weights are set as $\omega_1 = 5.0, \quad \omega_2 = 0.5,$ and $\quad \omega_3 = 1.0$.
The service weight reflects the primary mission objective, while exploration and coordination terms provide sufficient incentive for information gathering and efficient spatial distribution. Weight sensitivity analysis is included in Section 5.

\subsection{Problem Formulation}

We aim to maximize cumulative reward over a mission horizon of $T$ time steps by jointly optimizing agent trajectories $\mathcal{T} = \{\mathbf{p}_i(t)\}_{i,t}$:
\begin{subequations} 
\label{eq:problem} 
\begin{align} \max_{\mathcal{T}} \quad & \sum_{t=1}^{T} r_t, \label{eq:objective} \\ 
\text{s.t.} \quad & \|\mathbf{v}_i(t)\|_\infty \leq d_{\max}, \quad \forall i, \forall t, 
\label{eq:velocity_constraint} \\ & 
\mathbf{p}_i(t) \in \mathcal{A}, \quad \forall i, \forall t, \label{eq:boundary_constraint} \\ & 
\mathbf{p}_i(0) = \mathbf{p}_i^{\text{init}}, \quad \forall i, 
\label{eq:initial_constraint} \end{align} \end{subequations}
where Equation~\ref{eq:velocity_constraint} bounds the maximum displacement per step, Equation~\ref{eq:boundary_constraint} ensures agents remain within the service area, and Equation~\ref{eq:initial_constraint} specifies initial positions.

The optimization problem stated in Equation~\ref{eq:problem} is an instance of informative path planning (IPP) under partial observability. The underlying objective of maximizing information gain through sequential sensing is NP-hard in general~\citep{Krause2008Submodular}, and greedy approximations rely on submodularity that does not extend to the full reward structure. The addition of dynamic demand, continuous agent dynamics, and coordination between multiple mobile agents further precludes exact methods, motivating the use of reinforcement learning. However, as pure deep RL suffers from poor sample efficiency, we address this gap through HBRL framework that combines Bayesian spatial modeling with deep reinforcement learning via dual-channel knowledge transfer, detailed in Section~4.


\section{Proposed Solution}
\label{Proposed Solution}

\subsection{Two-Phase HBRL Framework}

We instantiate the proposed general HBRL framework of Section~\ref{System Model} for a multi-UAV wireless service provisioning scenario, where agents correspond to UAVs deployed as aerial base stations at fixed altitude, serving spatially distributed users. The spatial demand field corresponds to wireless service requests, the sensing footprint maps to each UAV's circular communication coverage of radius $r_c$, and the event probability $\pi_{\text{req}}$ corresponds to the per-user service request probability.

\begin{figure*}[!t]
    \centering
  \includegraphics[scale=1]{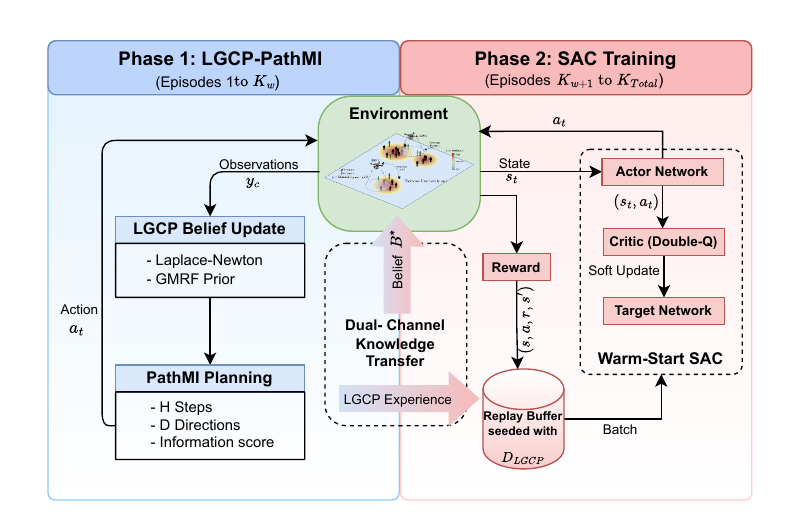}
  \caption{Overview of the proposed HBRL framework. Phase 1 employs LGCP-based belief inference and PathMI planning to guide information-driven exploration. Phase 2 performs policy optimization using Soft Actor-Critic (SAC), warm-started via dual-channel knowledge transfer: (i) belief state transfer and (ii) replay buffer seeding with LGCP-generated trajectories.}
  \label{Architecture}
\vspace{-4mm}
\end{figure*}
Figure~\ref{Architecture} summarizes the two-phases of HBRL framework. In Phase~1 (Exploration), UAVs execute information-driven trajectories guided by Pathwise Mutual Information (PathMI) planning while building spatial belief maps through LGCP inference. In Phase~2 (Exploitation), a Soft Actor-Critic (SAC) agent takes over trajectory control, initialized with knowledge transferred from Phase~1 through two complementary channels: (i) \textit{belief state transfer} via the trained LGCP belief maps informing the RL state representation, and (ii) \textit{behavioral transfer} via expert demonstrations from the LGCP phase seeding the RL replay buffer.

This hybrid design addresses a fundamental limitation of pure reinforcement learning approaches: the need to learn both \textit{where} high-value regions exist and \textit{how} to navigate efficiently. We separate these two questions by using LGCP for spatial learning and RL for policy optimization. The framework achieves superior sample efficiency compared to learning both from scratch.

\subsection{PathMI Information Score Planning}

During Phase~1, UAV trajectories are guided by an uncertainty-driven planner that evaluates candidate paths based on expected uncertainty reduction. In contrast to myopic (one-step) planners that focus on maximizing immediate information gain~\citep{Krause2007NonmyopicGP}, our approach considers multi-step lookahead to identify globally beneficial trajectories, following the receding-horizon paradigm common in model predictive control~\citep{Mayne2000MPC}. We use an uncertainty-weighted coverage score (incorporating variance and staleness) as a computationally efficient surrogate for mutual information, which we term PathMI.\footnote{A preliminary comparison of exploration strategies operating on the LGCP belief model is provided in Appendix~\ref{Appendix_A}.}

\subsubsection{Candidate Path Generation}
From each UAV's current grid position $(g_x, g_y)$, we generate $D_{dir} = 8$ candidate paths corresponding to the eight cardinal and diagonal directions. Each path extends for $L$ steps (the planning horizon), with positions computed as
\begin{equation}
    \mathbf{p}^{(\ell)} = \text{clip}(\mathbf{p}_i + \ell \cdot d_{\max} \cdot \mathbf{d}, \mathcal{A}),
    \label{eq:candidate_path}
\end{equation}
where $\mathbf{d} \in \{(\pm 1, 0), (0, \pm 1), (\pm 1, \pm 1)\}$ is the unit direction vector and $\ell \in \{1, \ldots, L\}$ indexes positions along the path. The resulting waypoints guide action selection, with actual displacement bounded by $d_{\max}$. The choice of straight-line paths balances computational efficiency against trajectory diversity; more complex path primitives could be substituted at increased computational cost.

\subsubsection{PathMI Score Computation}
For each candidate path $\mathcal{P}$, we compute the set of grid cells that would be observed along the trajectory:
\begin{equation}
\mathcal{O}(\mathcal{P}) = \bigcup_{\ell=1}^{L} \{c \in \mathcal{G} : \|\mathbf{p}_c - \mathbf{p}^{(\ell)}\|_2 \leq r_c\},
\label{eq:path_coverage}
\end{equation}
where $\mathbf{p}^{(\ell)}$ denotes the world-coordinate position of waypoint $\ell$ along path $\mathcal{P}$. The PathMI Information Score aggregates uncertainty over cells along the path:
\begin{equation}
    \text{PathMI}(\mathcal{P}) = \sum_{c \in \mathcal{O}(\mathcal{P})} \left[ \xi_c \cdot \sigma_c^2 + \omega_s \cdot \min\left(\frac{s_c}{s_{\max}}, 1\right) \cdot \sigma_c^2 \right],
    \label{eq:pathmi_score}
\end{equation}
where $\sigma_c^2$ is the current posterior variance, $s_c$ is the staleness (time steps since last observation) at cell $c$, $s_{\max}$ is a normalization constant equal to $T$, and $\omega_s$ is the staleness weighting coefficient. The staleness term encourages revisitation of regions where belief confidence has degraded. The factor $\xi_c$ down-weights cells with repeated observations, reflecting the diminishing returns characteristic of information gain~\citep{Krause2008Submodular}:
\begin{equation}
    \xi_c = \frac{1}{1 + n_c},
    \label{eq:xi_definition}
\end{equation}
where $n_c \geq 0$ denotes the cumulative observation count at cell $c$. Unvisited cells ($n_c = 0$) yield $\xi_c = 1$, providing maximum incentive for first-time exploration, while previously visited cells yield progressively smaller factors as observations accumulate. This formulation ensures that PathMI prioritizes novel regions while still permitting revisitation when staleness is high.

\textit{Remark 5:} The PathMI score (Equation~\ref{eq:pathmi_score}) approximates mutual information~\citep{Krause2007NonmyopicGP} through variance aggregation. Under Gaussian assumptions, high-variance regions yield greater entropy reduction upon observation, making cumulative variance a computationally efficient surrogate for ranking candidate paths by expected information gain~\citep{MacKay1992ActiveData,Riis2022BayesianActive}. True mutual information computation requires marginalization over future observations, which is computationally prohibitive for online planning.\footnote{The sequence of approximations relating Equation~\ref{eq:pathmi_score} to the expected information gain is made explicit in Appendix~\ref{Appendix_E}}.

\subsubsection{PathMI Action Selection}
The UAV agent selects the path with maximum PathMI score:
\begin{equation}
    \mathcal{P}^* = \arg\max_{\mathcal{P} \in \{\mathcal{P}_1, \ldots, \mathcal{P}_D\}} \text{PathMI}(\mathcal{P}).
    \label{eq:best_path}
\end{equation}
The action is then computed as the normalized direction toward the first waypoint of $\mathcal{P}^*$:
\begin{equation}
    \mathbf{a}_i = \frac{\Delta \mathbf{p}}{\max\left(\|\Delta \mathbf{p}\|_2, \, \epsilon\right)},
    \label{eq:pathmi_action}
\end{equation}
where $\Delta \mathbf{p} = \mathbf{p}_{\mathcal{P}^*(1)} - \mathbf{p}_i$ denotes the displacement to the first waypoint, $\mathbf{p}_{\mathcal{P}^*(1)}$ is the world position of that waypoint, and $\epsilon > 0$ is a small constant (e.g., $\epsilon = 10^{-6}$) preventing division by zero when the waypoint coincides with the current position after clipping. The action is then clipped to $[-1, 1]^2$ before execution. This receding-horizon approach re-plans at each step, adapting to updated beliefs while maintaining non-myopic trajectory structure.

\subsection{Dual-Channel Knowledge Transfer}

 In contrast to prior approaches that transfer either state representations~\citep{Zhu2023TransferRL} or behavioral demonstrations~\citep{Hester2018DQfD}, but not both simultaneously, our framework transfers both channels, enabling complementary benefits.

\subsubsection{Channel 1: Belief State Transfer}
At the conclusion of Phase~1, the LGCP belief state $\mathcal{B}^* = \{\boldsymbol{\mu}_u^*, \boldsymbol{\sigma}^{2*}\}$ encodes the spatial demand information learned during information-driven exploration. This belief is transferred to Phase~2 by initializing the reinforcement learning environment at the start of training with $\mathcal{B}_{\text{RL}}(0) \leftarrow \mathcal{B}^*$.
Belief state transfer serves as an initialization mechanism that provides an informed prior for early policy gradient updates, mitigating instability associated with uninformed exploration during the initial stages of training. To preserve episodic independence and avoid cross-episode accumulation of spatial information, subsequent episodes reset the belief state to the uninformative prior $\mathcal{B}_0 = \{\mathbf{0}, \mathbf{1}\}$. Learning progress is retained in the policy parameters and replay buffer rather than in the belief state itself.

\subsubsection{Channel 2: Behavioral Transfer}
During Phase~1, the LGCP planner generates high-quality trajectories optimized for information acquisition. We store these trajectories as demonstration data:
\begin{equation}
    \mathcal{D}_{\text{LGCP}} = \{(\mathbf{s}_t, \mathbf{a}_t, r_t, \mathbf{s}_{t+1}, d_t)\}_{t=1}^{T_w},
    \label{eq:demo_buffer}
\end{equation}
where $T_w = K_w \times T$ is the total number of transitions collected over $K_w$ warm-start episodes, and $d_t \in \{0, 1\}$ is the terminal flag indicating whether $\mathbf{s}_{t+1}$ is a terminal state. Following the learning from demonstrations paradigm~\citep{Hester2018DQfD}, these transitions seed the SAC replay buffer before training begins:
\begin{equation}
    \mathcal{D}_{\text{SAC}} \leftarrow \mathcal{D}_{\text{LGCP}}.
    \label{eq:buffer_seeding}
\end{equation}
This seeding provides two benefits: (i) the agent can immediately sample informative transitions for gradient updates without random exploration, and (ii) the demonstrations encode implicit coordination patterns that accelerate multi-UAV coordination learning.

The two channels address different aspects of the learning problem. The belief channel provides uncertainty estimates for reward shaping (overlap penalty) and global belief summaries indicating overall exploration progress. The behavioral channel provides state-action pairs demonstrating effective spatial exploration patterns. Ablation studies in Section~\ref{sec:results} quantify the individual and combined contributions of each channel, demonstrating that full dual-channel transfer outperforms either channel alone.

The complementary effect can be interpreted through early state-distribution alignment and reward calibration. The replay buffer $\mathcal{D}_{\mathrm{LGCP}}$ contains transitions collected under the evolving Phase-1 belief, where the state summary $\boldsymbol{\phi} = [\bar{\lambda}, \bar{\sigma}^2, \bar{n}_{\text{obs}}]^\top$ (cf. Equation~\ref{eq:state_space}) reflects structured spatial information. If SAC instead begins from the uninformed prior $\mathcal{B}_0$, the initial on-policy state distribution over $\boldsymbol{\phi}$ differs from that of the buffered demonstrations, creating a distribution mismatch during early critic updates. Initializing the first SAC episode with $\mathcal{B}^*$ can reduce this mismatch, improving alignment between replayed and current transitions at the start of training. Moreover, both the exploration reward and the variance-normalized overlap penalty depend on calibrated posterior uncertainty. Under $\mathcal{B}_0$, posterior variance is initially uniform, reducing reward contrast across actions. Initializing with $\mathcal{B}^*$ restores spatial structure at the start of Phase 2, strengthening reward contrast across action.

\textit{Remark 6 (Distribution mismatch impacts off-policy estimates):} Let $d_{\pi}(s)$ be the discounted state visitation distribution under policy $\pi$ and $d_{\mathcal{D}}(s)$ the empirical state distribution induced by samples from replay buffer $\mathcal{D}$. For any bounded function $f:\mathcal{S}\rightarrow \mathbb{R}$,
\begin{equation}
\left|
\mathbb{E}_{s\sim d_{\pi}}[f(s)] - \mathbb{E}_{s\sim d_{\mathcal{D}}}[f(s)]
\right|
\le \|f\|_{\infty}\,\mathrm{V}(d_{\pi}, d_{\mathcal{D}}),
\end{equation}
where $\mathrm{V}(\cdot,\cdot)$ denotes total variation distance. This follows directly from the definition of total variation distance, $\mathrm{V}(P,Q) = \sup_A |P(A) - Q(A)|$. Hence, when replay and on-policy state distributions differ, quantities learned under $d_{\mathcal{D}}$ (e.g., value estimates) may generalize poorly to states encountered under $d_{\pi}$.

In our setting, $\mathcal{D}_{\text{LGCP}}$ contains transitions collected under belief $\mathcal{B}^*$, inducing a structured distribution over the state summary $\boldsymbol{\phi}$. Initializing the first SAC episode with $\mathcal{B}^*$ reduces the initial mismatch relative to starting from $\mathcal{B}_0$, thereby improving the relevance of early critic updates.

\subsubsection{Episode Reset Policy}

At the start of each SAC episode in Phase~2, the LGCP belief state is reset to the uninformed prior $\mathcal{B}_0 = \{\mathbf{0}, \mathbf{1}\}$. This prevents accumulation of spatial information across episodes, ensuring that each episode begins from an unknown demand state. Learning progress is retained in the policy parameters and replay buffer, not in the belief state. The transferred belief $\mathcal{B}^*$ is used only to initialize the \textit{first} SAC episode, providing a non-trivial starting point for initial policy gradient updates.

\subsection{MDP Formulation}
\label{sec:mdp}

The multi-UAV coordination problem is a partially observable Markov decision process (POMDP), as agents observe demand only within their sensing footprint. Following the belief-space Markov decision process (MDP) formulation~\citep{kaelbling1998planning}, we train the policy on compressed observations. Compact belief features in place of the full belief follow established belief-compression approaches in POMDP planning (\cite{roy1999coastal}, \cite{roy2005finding}) and active-sensing RL (\cite{jeong2019learning}, \cite{subramaniam2025track}). The unknown demand field is tracked internally through LGCP-based belief updates, and the policy receives low-dimensional belief summaries as state input. The environment maintains the full belief for state evolution and reward computation. Rationale and implications are discussed below in Remark 7.

\subsubsection{State Space}
The state vector $\mathbf{s}_t \in \mathbb{R}^{3N+4}$ comprises:
\begin{equation}
    \mathbf{s}_t = \left[ \tilde{\mathbf{p}}_1, \ldots, \tilde{\mathbf{p}}_N, \, \boldsymbol{\phi}, \, \bar{c}_1, \ldots, \bar{c}_N, \, \frac{t}{T} \right]^\top,
    \label{eq:state_space}
\end{equation}
where $\tilde{\mathbf{p}}_i = 2(\mathbf{p}_i / [L_x, L_y]^\top) - \mathbf{1} \in [-1, 1]^2$ denotes the normalized UAV positions ($2N$ dimensions), $\boldsymbol{\phi} = [\bar{\lambda}, \bar{\sigma}^2, \bar{n}_{\text{obs}}]^\top \in \mathbb{R}^3$ represents the global belief summary and is computed as arithmetic mean intensity, mean variance, mean observation count over the operational regions assigned to the UAVs, rather than over the entire grid, $\bar{c}_i \in [0, 1]$ indicates coverage progress for UAV $i$ ($N$ dimensions), and $t/T \in [0, 1]$ is the normalized time step.

The coverage progress $\bar{c}_i$ for UAV $i$ is computed as the fraction of grid cells visited by UAV $i$ up to time $t$:

\begin{equation}
    \bar{c}_i(t)
    =
    \frac{1}{|\mathcal{G}_i|}
    \sum_{c\in\mathcal{G}_i}
    \mathbb{I}\!\left(
    \exists\, t' \le t : c\in\mathcal{C}_i(t')
    \right)
    \label{eq:coverage_progress}
\end{equation}

\textit{Remark 7:} We intentionally use compact global belief summaries rather than the full belief grid to keep the state low-dimensional and scalable. While this compression sacrifices local spatial detail (the agent knows ``overall uncertainty is high'' but not ``where''), it ensures stable training and $\mathcal{O}(1)$ state growth with grid size. This design reflects a deliberate trade-off between representational fidelity and scalability. The UAV positions provide explicit spatial grounding. Empirical results in Section \ref{Learning Performance and Efficiency} confirm that the compressed state retains decision-relevant structure for stable and improved performance.

\subsubsection{Action Space}
The joint action $\mathbf{a}_t \in [-1, 1]^{2N}$ specifies bounded motion commands for all UAVs:
\begin{equation}
    \mathbf{a}_t = [\mathbf{a}_1^\top, \ldots, \mathbf{a}_N^\top]^\top, \quad \mathbf{a}_i \in [-1, 1]^2.
    \label{eq:action_space}
\end{equation}
Each component represents a normalized displacement direction, mapped to per-step motion via $\mathbf{v}_i = \mathbf{a}_i \cdot d_{\max}$ as defined in Equation~\ref{eq:action_mapping}.

\subsubsection{Reward Function}
The reward $r_t$ follows the three-objective structure defined in Equation~\ref{eq:reward}, balancing service quality ($R_{\text{service}}$), exploration incentive ($R_{\text{explore}}$), and coordination efficiency ($C_{\text{coord}}$). In the wireless service provisioning context, these terms map to established operational objectives: $R_{\text{service}}$ (total served demand) aligns with throughput optimisation~\citep{Wu2018JointTrajectoryComm} and user quality of service~\citep{Hao2024TaskOffloadingPriority}; $R_{\text{explore}}$ (belief improvement) aligns with reducing demand uncertainty, a standard objective in information-driven planning~\citep{Krause2008Submodular}; and $C_{\text{coord}}$ (overlap and travel) aligns with spectrum-contention avoidance~\citep{rizvi2024multi} and UAV energy budget~\citep{Zeng2019EnergyMinimization}. Following~\cite{rizvi2024multi}, we deliberately abstract from application-specific models (wireless channel modelling, UAV propulsion energy, flight dynamics) to focus on the framework-level contribution.

\subsubsection{Transition Dynamics}
State transitions follow the deterministic dynamics in Equation~\ref{eq:position_update}, with stochasticity arising from the binomial observation model (Equation~\ref{eq:observation_model}) and the LGCP belief update.

\subsection{Soft Actor-Critic Training}

The continuous $2N$-dimensional action space precludes discrete-action methods such as DQN~\citep{Mnih2015DQN}. Among continuous-action algorithms, we select Soft Actor-Critic (SAC)~\citep{Haarnoja2018SAC} for three reasons: (i) its off-policy nature enables replay buffer seeding with LGCP demonstrations, whereas on-policy methods like Proximal Policy Optimization (PPO)~\citep{Schulman2017PPO} would discard these transitions; (ii) entropy regularization maintains exploration during the transition from demonstration-driven to self-generated experience, preventing premature convergence; and (iii) clipped double Q-learning provides greater stability than Deep Deterministic Policy Gradient (DDPG)~\citep{Lillicrap2016DDPG} in joint action settings for multiple agents.

\begin{table*}[!t]
\caption{Simulation and Algorithm Parameters}
\label{tab:parameters}
\begin{center}
\begin{tabular}{ll@{\hspace{1.2cm}}ll}
\textbf{Parameter} & \textbf{Value} &
\textbf{Parameter} & \textbf{Value} \\
\hline \\[-0.9em]
Service area ($\mathcal{A}$) & $2000 \times 2000$ m$^2$ &
Hidden layers & $256$--$256$ \\

Grid resolution ($\Delta$) & $20$ m &
Learning rate ($\eta$) & $3 \times 10^{-4}$ \\

Grid size ($G_x \times G_y$) & $100 \times 100$ &
Replay buffer size ($|\mathcal{D}|$) & $10^4$ \\

UAVs ($N$) & $2-4$ &
Batch size ($B$) & $256$ \\

Coverage radius ($r_c$) & $100$ m &
Discount factor ($\gamma$) & $0.99$ \\

Max displacement ($d_{\max}$) & $15$ m &
Soft update ($\tau_{\text{soft}}$) & $0.005$ \\

No. of Hotspots ($J$) & $3$--$5$ &
Target entropy ($\bar{\mathcal{H}}$) & $-\dim(\mathcal{A})$ \\

Hotspot Diffusion rate ($\sigma_h$) & $1$ m/step &
Learning starts & $100$ steps \\

Displacement bound ($r_b$) & $75$ m &
Total episodes ($K$) & $200$ \\

Request prob.\ ($\pi_{\text{req}}$) & $0.05$ &
Warm-start episodes ($K_w$) & $10$--$50$ \\

Smoothness ($\beta$) & $0.2$ &
Episode length ($T$) & $200$ \\

Variance growth ($\gamma_v$) & $0.002$ &
Planning horizon ($L$) & $5$ \\

Variance bounds ($\sigma^2_{\min/\max}$) & $0.01$ / $1.0$ &
Candidate directions ($D_{dir}$) & $8$ \\

Staleness weight ($\omega_s$) & $0.1$ &
Service weight ($\omega_1$) & $5.0$ \\

Prior precision ($\tau$) & $1.0$ &
Exploration weight ($\omega_2$) & $0.5$ \\

Sampling trials ($n_s$) & $100$ &
Coordination weight ($\omega_3$) & $1.0$ \\

Newton iterations ($n_{\text{Newton}}$) & $3$ &
PCG iterations ($n_{\text{PCG}}$) & $8$ \\

Max staleness ($s_{\max}$) & $T$ & \\

\end{tabular}
\end{center}
\end{table*}

\subsubsection{Maximum Entropy Objective}
SAC optimizes a maximum entropy objective~\citep{Haarnoja2018SAC}:
\begin{equation}
    J(\pi) = \mathbb{E}_{\tau \sim \pi} \left[ \sum_{t=0}^{T} \gamma^t \left( r(\mathbf{s}_t, \mathbf{a}_t) + \alpha \mathcal{H}(\pi(\cdot | \mathbf{s}_t)) \right) \right],
    \label{eq:sac_objective}
\end{equation}
where $\gamma \in (0, 1)$ denotes the discount factor, $\alpha > 0$ is the temperature parameter controlling exploration-exploitation balance, and $\mathcal{H}(\cdot)$ represents entropy. The entropy regularization encourages exploration during early training while allowing convergence to near-deterministic policies as learning progresses.

\subsubsection{SAC Architecture}
Following standard practice~\citep{Haarnoja2018SAC}, the policy and value networks use multi-layer perceptrons (MLPs) with two hidden layers of 256 units each and ReLU activations. The policy network $\pi_\theta$ outputs mean $\boldsymbol{\mu}_\theta(\mathbf{s})$ and log-standard-deviation $\log \boldsymbol{\sigma}_\theta(\mathbf{s})$ for a squashed Gaussian distribution:
\begin{equation}
    \mathbf{a} = \tanh\left(\boldsymbol{\mu}_\theta(\mathbf{s}) + \boldsymbol{\sigma}_\theta(\mathbf{s}) \odot \boldsymbol{\epsilon}\right), \quad \boldsymbol{\epsilon} \sim \mathcal{N}(\mathbf{0}, \mathbf{I}),
    \label{eq:squashed_gaussian}
\end{equation}
where $\tanh(\cdot)$ bounds actions to $[-1, 1]^{2N}$. Two Q-networks $Q_{\phi_1}, Q_{\phi_2}$ with separate target networks $Q_{\bar{\phi}_1}, Q_{\bar{\phi}_2}$ mitigate overestimation bias through clipped double Q-learning~\citep{Fujimoto2018TD3}.

\subsubsection{Loss Functions}
The Q-networks are trained by minimizing the soft Bellman residual:
\begin{equation}
    \mathcal{L}_Q(\phi_j) = \mathbb{E}_{(\mathbf{s}, \mathbf{a}, r, \mathbf{s}', d) \sim \mathcal{D}} \left[ \left( Q_{\phi_j}(\mathbf{s}, \mathbf{a}) - y \right)^2 \right],
    \label{eq:q_loss}
\end{equation}
where the target $y$ is computed as:
\begin{equation}
    y = r + \gamma (1 - d) \left( \min_{j=1,2} Q_{\bar{\phi}_j}(\mathbf{s}', \tilde{\mathbf{a}}') - \alpha \log \pi_\theta(\tilde{\mathbf{a}}' | \mathbf{s}') \right),
    \label{eq:q_target}
\end{equation}
with $\tilde{\mathbf{a}}' \sim \pi_\theta(\cdot | \mathbf{s}')$ and $d \in \{0, 1\}$ indicating terminal states. The policy is updated by maximizing the expected return with entropy bonus:
\begin{equation}
    \mathcal{L}_\pi(\theta) = \mathbb{E}_{\mathbf{s} \sim \mathcal{D}, \tilde{\mathbf{a}} \sim \pi_\theta} \left[ \alpha \log \pi_\theta(\tilde{\mathbf{a}} | \mathbf{s}) - \min_{j=1,2} Q_{\phi_j}(\mathbf{s}, \tilde{\mathbf{a}}) \right].
    \label{eq:policy_loss}
\end{equation}
The temperature $\alpha$ is automatically tuned to maintain a target entropy $\bar{\mathcal{H}}$~\citep{Haarnoja2018SACv2}:
\begin{equation}
    \mathcal{L}_\alpha = \mathbb{E}_{\tilde{\mathbf{a}} \sim \pi_\theta} \left[ -\alpha \left( \log \pi_\theta(\tilde{\mathbf{a}} | \mathbf{s}) + \bar{\mathcal{H}} \right) \right].
    \label{eq:alpha_loss}
\end{equation}

\subsubsection{Target Network Updates}
Target networks are updated via exponential moving average (Polyak averaging)~\citep{Lillicrap2016DDPG}:
\begin{equation}
    \bar{\phi}_j \leftarrow \tau_{\text{soft}} \phi_j + (1 - \tau_{\text{soft}}) \bar{\phi}_j,
    \label{eq:target_update}
\end{equation}
where $\tau_{\text{soft}} \ll 1$ ensures stable target values.

\begin{figure*}[!t]
\centering

\begin{minipage}[t]{0.485\textwidth}
\footnotesize
\textbf{Algorithm 1: Phase 1 -- LGCP--PathMI Exploration}
\vspace{0.3em}
\hrule
\vspace{0.3em}

\begin{algorithmic}[1]
\State Initialize log-intensity $\mathbf{u} \leftarrow \mathbf{0}$, variance $\boldsymbol{\sigma}^2 \leftarrow \mathbf{1}$
\State Initialize observation counts $\mathbf{n} \leftarrow \mathbf{0}$, staleness $\mathbf{s} \leftarrow \mathbf{0}$
\State Initialize precision matrix $\mathbf{Q} = \tau \mathbf{I} + \beta \mathbf{L}_G$
\State Initialize demonstration buffer $\mathcal{D}_{\text{LGCP}} \leftarrow \emptyset$

\For{episode $k = 1$ to $K_w$}
    \State Reset environment; reset belief to prior $\mathcal{B}_0$
    \State Reset staleness $\mathbf{s} \leftarrow 0$, observation counts $\mathbf{n} \leftarrow 0$
        \For{$t = 0$ to $T-1$}
        \State Observe state $\mathbf{s}_t$
        \State Update staleness: $s_c \leftarrow s_c + 1$ for all $c \in \mathcal{G}$
        \For{each UAV $i = 1$ to $N$}
            \State Generate $D_{dir}$ candidate paths Equation~\ref{eq:candidate_path}
            \State Compute PathMI scores Equation~\ref{eq:pathmi_score}
            \State Select $\mathcal{P}^* = \arg\max_d \text{PathMI}(\mathcal{P}_d)$
            \State Set $\mathbf{a}_i$ toward first waypoint Equation~\ref{eq:pathmi_action}
        \EndFor
        \State Execute joint action $\mathbf{a}_t$, observe $\mathbf{y}(t)$
        \State For all $c \in \mathcal{C}(t)$: $e_c \leftarrow 1$, \; $y_c \leftarrow y_c(t)$, \; $n_c \leftarrow n_c + 1$
        \State Apply variance growth Equation~\ref{eq:variance_predict}; store $\tilde{\sigma}_c^2(t)$
        \State Update LGCP belief Equation~\ref{eq:newton_update}
        \State Apply variance update Equation~\ref{eq:variance_update} $\rightarrow \sigma_c^2(t^+)$
        \State Compute reward $r_t$ Equation~\ref{eq:reward} using $\tilde{\sigma}_c^2(t)$ and $\sigma_c^2(t^+)$
        \State Compute next state $\mathbf{s}_{t+1}$
        \State Set terminal flag $d_t \leftarrow \mathbb{I}(t = T-1)$
        \State Reset staleness: $s_c \leftarrow 0$ for all $c \in \mathcal{C}(t)$
        \State Store $(\mathbf{s}_t, \mathbf{a}_t, r_t, \mathbf{s}_{t+1}, d_t)$ in $\mathcal{D}_{\text{LGCP}}$
    \EndFor
\EndFor
\State Set final belief $\mathcal{B}^* = \{\boldsymbol{\mu}_u, \boldsymbol{\sigma}^2\}$\\
\Return $\mathcal{B}^*$, $\mathcal{D}_{\text{LGCP}}$
\end{algorithmic}

\vspace{0.3em}
\hrule
\end{minipage}
\hfill
\begin{minipage}[t]{0.485\textwidth}
\footnotesize
\textbf{Algorithm 2: Phase 2 -- Warm-Started SAC Training}
\vspace{0.3em}
\hrule
\vspace{0.3em}

\begin{algorithmic}[1]
\State Initialize replay buffer $\mathcal{D} \leftarrow \mathcal{D}_{\text{LGCP}}$ \hfill $\triangleright$ Buffer seeding
\State Initialize actor $\pi_\theta$, critics $Q_{\phi_1}, Q_{\phi_2}$
\State Initialize targets $\bar{Q}_{\phi_1} \leftarrow Q_{\phi_1}$, $\bar{Q}_{\phi_2} \leftarrow Q_{\phi_2}$
\State Initialize temperature $\alpha$

\For{episode $k = K_w + 1$ to $K$}
    \If{$k = K_w + 1$}
        \State Initialize belief with $\mathcal{B}^*$ from Phase 1
    \Else
        \State Reset belief to prior $\mathcal{B}_0 = \{\mathbf{0}, \mathbf{1}\}$
    \EndIf
    \State Reset staleness $\mathbf{s} \leftarrow \mathbf{0}$, observation counts $\mathbf{n} \leftarrow \mathbf{0}$
    \For{$t = 0$ to $T-1$}
        \State Observe $\mathbf{s}_t$, sample $\mathbf{a}_t \sim \pi_\theta(\cdot | \mathbf{s}_t)$
        \State Execute $\mathbf{a}_t$, observe $y(t)$
        \State For all $c \in \mathcal{C}(t)$: $e_c \leftarrow 1$, \; $y_c \leftarrow y_c(t)$, \; $n_c \leftarrow n_c + 1$
        \State Apply variance growth Equation~\ref{eq:variance_predict} $\rightarrow \tilde{\sigma}_c^2(t)$
        \State Update LGCP belief via Equation~\ref{eq:newton_update}
        \State Apply variance update Equation~\ref{eq:variance_update} $\rightarrow \sigma_c^2(t^+)$
        \State Compute reward $r_t$ via Equation~\ref{eq:reward}
        \State Compute next state $\mathbf{s}_{t+1}$
        \State Set terminal flag $d_t \leftarrow \mathbb{I}(t = T-1)$
        \State Store $(\mathbf{s}_t, \mathbf{a}_t, r_t, \mathbf{s}_{t+1}, d_t)$ in $\mathcal{D}$
        \State Sample mini-batch from $\mathcal{D}$
        \State Update critics and actor via Equation \ref{eq:q_loss} \& \ref{eq:policy_loss} 
        \State Update temperature $\alpha$ via Equation \ref{eq:alpha_loss}
        \State Soft-update target networks via Equation \ref{eq:target_update}
    \EndFor
\EndFor\\
\Return trained policy $\pi_\theta$
\end{algorithmic}

\vspace{0.3em}
\hrule
\end{minipage}

\caption{Two-phase training pipeline: LGCP--PathMI exploration (left) and warm-started SAC training (right).}
\label{two_algorithms}
\end{figure*}
The complete framework is presented in Figure~\ref{two_algorithms}. Algorithm 1 details the LGCP-PathMI exploration phase, while Algorithm 2 describes the warm-started SAC training phase.

\textit{Remark 8:} Integration of LGCP with reinforcement learning for spatial exploration and service provisioning is less explored. Our \textit{methodological} contribution lies in the dual-channel transfer mechanism and the variance-normalized coordination penalties, building upon the LGCP or SAC components.

\subsection{Complexity Analysis}

\begin{table}[!t]
\centering
\caption{Wall-clock computational cost of the principal HBRL components under the standard experimental configuration shows the belief-guided warm-start phase has a lower per-step runtime than subsequent Phase 2 training in the present implementation.}
\label{tab:runtime_overhead}
\begin{tabular}{lcc}
\hline
\textbf{Component} & \textbf{Mean $\pm$ SD (ms)} & \textbf{Median (ms)} \\
\hline
LGCP belief update & $4.33 \pm 0.58$ & $4.27$ \\
PathMI planning & $23.54 \pm 0.90$ & $23.41$ \\
Phase 1 (LGCP--PathMI) & $27.86 \pm 1.17$ & $27.73$ \\
Phase 2 (LGCP--SAC) & $37.45 \pm 4.50$ & $34.92$ \\
\hline
\end{tabular}
\end{table}
\subsubsection{Time Complexity}
The LGCP belief update via Laplace-Newton requires $\mathcal{O}(n_{\text{Newton}} \cdot n_{\text{PCG}} \cdot |\mathcal{G}|)$ operations per step, where $n_{\text{Newton}}$ and $n_{\text{PCG}}$ denote the number of Newton and PCG iterations, respectively, and $|\mathcal{G}| = G_x \times G_y$ is the union of operational grid size. PathMI planning evaluates $D_{dir}$ candidate paths, each covering $\mathcal{O}(L \cdot r_g^2)$ cells where $r_g = \lceil r_c / \Delta \rceil$ is the coverage radius in grid cells, yielding $\mathcal{O}(D \cdot L \cdot r_g^2)$ per UAV per step. SAC updates are $\mathcal{O}(B \cdot d_{\text{net}})$ where $B$ is the batch size and $d_{\text{net}}$ denotes the network parameter count. To complement the asymptotic complexity analysis, we empirically measure the wall-clock computational cost of the principal HBRL components under the standard experimental configuration (Table \ref{tab:runtime_overhead}).

\subsubsection{Space Complexity}
The LGCP belief requires $\mathcal{O}(|\mathcal{G}|)$ storage for mean, variance, and exposure fields. The replay buffer stores $\mathcal{O}(|\mathcal{D}| \cdot d_s)$ transitions where $d_s = 3N + 4$ is the state dimension. SAC networks require $\mathcal{O}(d_{\text{net}})$ parameters.

\subsubsection{Scalability}
The framework scales linearly with the number of UAVs $N$ in state dimension ($3N + 4$) and action dimension ($2N$). For moderate fleet sizes, the joint policy formulation provides sample-efficient implicit coordination through shared belief state. Learning by parameter-sharing or other techniques for very large fleet sizes remains an open challenge for future work. For very large grids, the GMRF structure ensures sparse precision matrices, maintaining computational tractability.


\section{Experimental Evaluation}
\label{sec:results}

\subsection{Experimental Setup}
\label{Experimental Setup}

We perform simulations on an Apple M1 Pro with 3.2 GHz CPU, 16 GB RAM, and 16-core GPU, programmed in Python using Stable-Baselines3 for SAC implementation. The service area is $2000 \times 2000$ m$^2$ discretized into grids with resolution $\Delta = 20$ m, with $N=2$ UAVs operating at fixed altitude with a coverage radius $r_c = 100$ m. $r_c = 5\Delta$ ensures that each UAV footprint spans multiple grid cells and avoids discretization artefacts observed when $r_c < \Delta$. User demand is generated as a mixture of $J=3-5$ Gaussian hotspots with randomly sampled centers and spreads. All results are averaged over 5 independent seeds. Shaded bands and error bars show the range and standard deviation respectively across the seeds. We compare the proposed HBRL against the following baselines: Pure LGCP-PathMI (no RL), Pure SAC (cold start), BC+SAC (imitation warm start) and two ablations: SAC + Belief Transfer Only, and SAC + Buffer Seeding Only. All methods use identical SAC parameters summarized in Table~\ref{tab:parameters}. Pure SAC with the identical policy architecture and learning configuration serves as the principal RL baseline, allowing the effect of the proposed belief-guided warm start to be isolated without confounding it with differences in the underlying RL architecture. Values for SAC hyper-parameters follow the standard recommendations from the original SAC implementation~\citep{Haarnoja2018SAC}. All methods are evaluated under the same budget of $K=200$ episodes, i.e.\ $K \times T$ environment interactions. For HBRL the first $K_w$ episodes constitute the LGCP--PathMI warm start and SAC is trained for the remaining $K-K_w$ episodes, so the warm-start interactions are drawn from, and not added to, this budget. Unless stated otherwise, default parameters are used throughout.

As an imitation-based warm-start baseline, we include Behavior Cloning (BC) followed by reinforcement learning. After Phase 1 the SAC actor is first pre-trained offline via supervised learning to minimize negative log-likelihood over Phase~1 PathMI state-action pairs, using the same compressed state representation as the proposed method. After pre-training, the imitation objective is removed and the policy is fine-tuned using standard SAC updates with an empty replay buffer. This baseline uses identical Phase~1 data as our approach but incorporates it through actor pre-training rather than replay buffer seeding, allowing direct comparison of demonstration utilization strategies.

We evaluate our proposed framework through a series of experiments designed to assess learning efficiency, robustness, and scalability. Section~\ref{Learning Performance and Efficiency} compares learning performance against baselines. Sections~\ref{Warm start and Planning Horizon Ablations} and \ref{Scalability and coordination analysis} examine the sensitivity to warm-start duration, planning horizon, and fleet size. Section~\ref{Belief Dynamics and unvertainty Calibration} analyzes belief dynamics and uncertainty calibration, and Section~\ref{robustness to observation loss} examines robustness to experience loss.

\subsection{Learning Performance and Efficiency}
\label{Learning Performance and Efficiency}

\begin{figure}[!t]
\centering

\begin{minipage}[t]{0.48\textwidth}
    \centering
    \includegraphics[width=\linewidth]{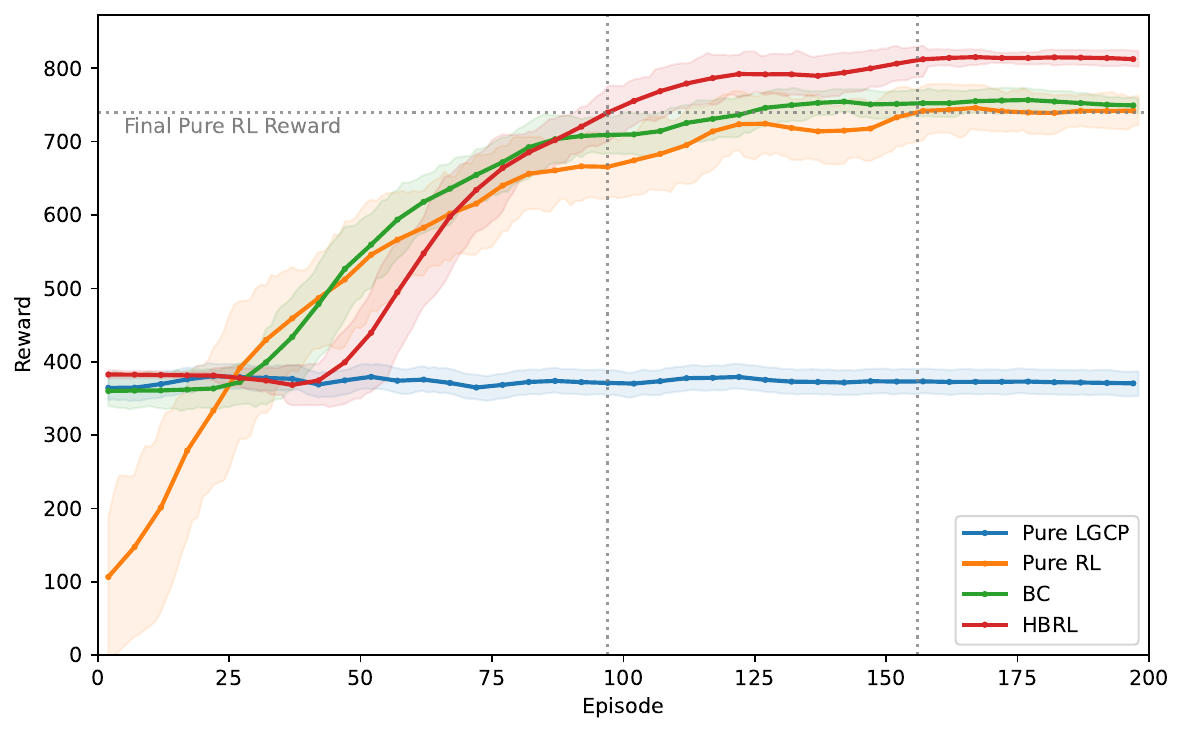}
    \captionof{figure}{Reward comparison between Pure LGCP, Pure RL, Behavior Cloning and HBRL frameworks.}
    \label{reward_vs_technique_block}
\end{minipage}
\hfill
\begin{minipage}[t]{0.48\textwidth}
    \centering
    \includegraphics[width=\linewidth]{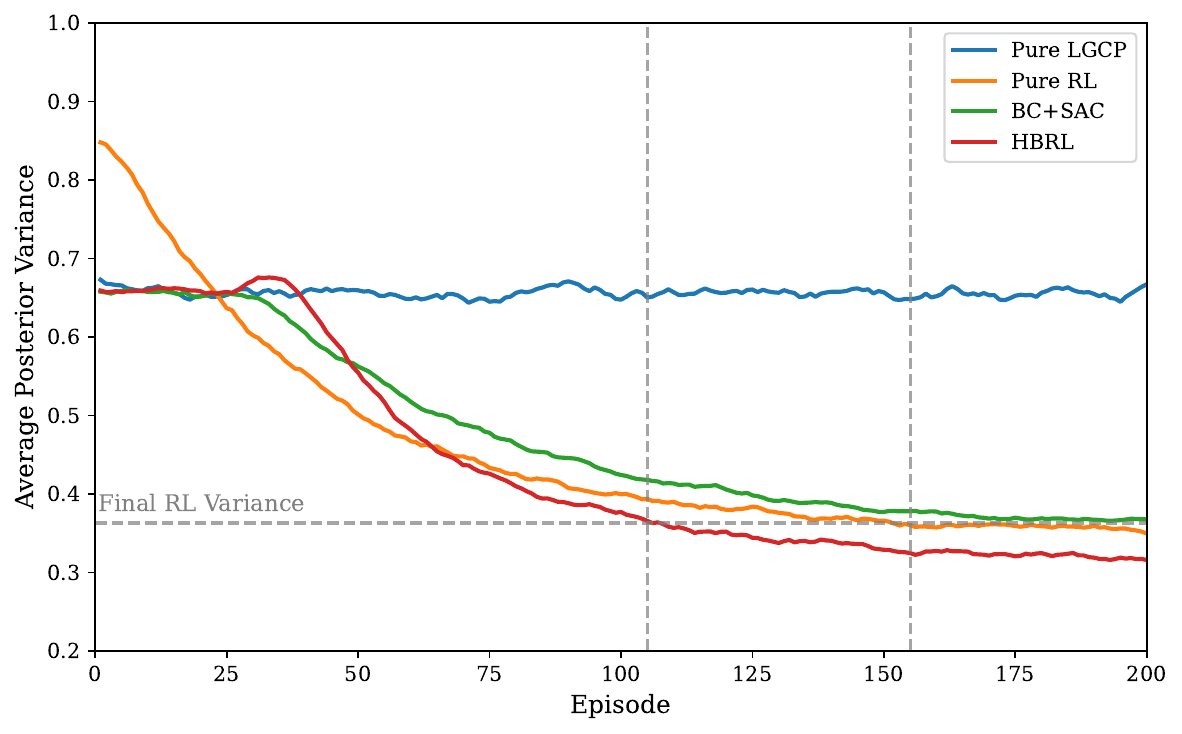}
    \captionof{figure}{Posterior variance comparison between Pure LGCP, Pure RL and HBRL. Lower values indicate higher confidence in the inferred demand field.}
    \label{variance_comparison}
\end{minipage}

\end{figure}

Figure~\ref{reward_vs_technique_block} compares learning behaviour across all methods. Pure LGCP maintains stable reward without policy adaptation, while Pure RL begins with the lowest reward due to random exploration before gradually improving. BC follows the LGCP reward levels during Phase~1 and transitions smoothly without the dip exhibited by HBRL, yet converges to similar asymptotic reward as Pure RL, indicating that action imitation accelerates early learning but does not improve final policy quality. HBRL eventually outperforms both Pure RL and BC+SAC, improving over Pure RL by 10.8\%, and reaching its final reward level 38\% earlier. Despite the compressed belief, learning through HBRL remains stable and outperforms both baselines, suggesting the summary retains relevant structure. The gap between BC+SAC and HBRL demonstrates that replay buffer seeding provides benefits beyond behavioral imitation.


While reward curves capture overall task performance, it is equally important to examine how each method influences the quality of the learned belief state, since accurate estimation of user demand is fundamental to the requirements of service providers. Figure~\ref{variance_comparison} shows the evolution of average posterior variance (diagonal Laplace proxy), where lower values indicate higher confidence in the inferred demand field. Pure LGCP maintains a nearly constant variance of approximately 0.66 because each episode begins with a fresh belief map, preventing long-term accumulation of information. Pure RL, in contrast, steadily reduces variance over time, converging to 0.37 by episode 155 as the agent revisits informative regions during policy improvement. HBRL, whereby starting learning later than Pure-RL, achieves the fastest and best variance reduction, matching best Pure-RL performance 32\% faster. It also reduces the variance against the benchmark by 11\%.The corresponding learned intensity maps are shown in Appendix~\ref{Appendix_B}.

\begin{figure*}[!t]
    \centering
    \includegraphics[width=4in]{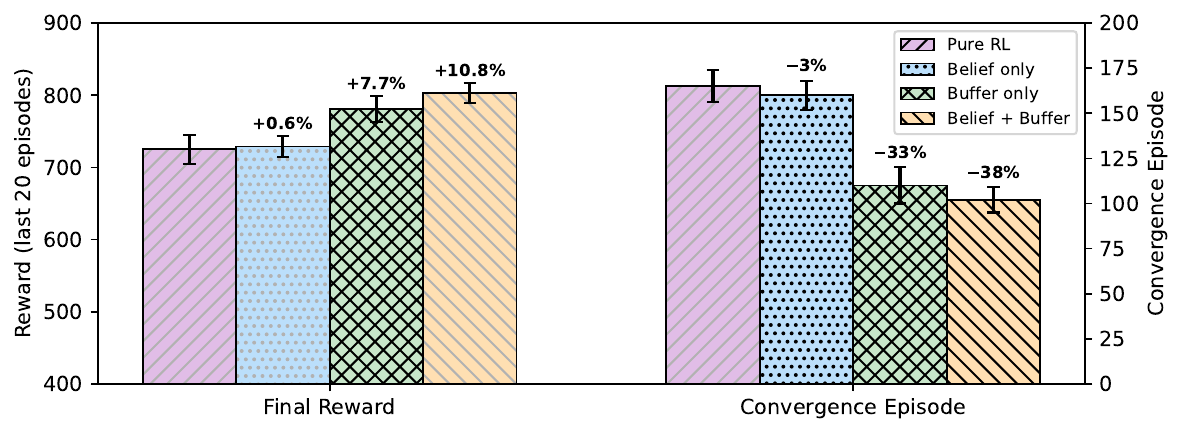}
    \caption{Comparison of reward and episodes to reach Pure RL reward for all three       
  transfer channel scenarios.}\label{transfer_channel_ablation}
\end{figure*}
Figure~\ref{transfer_channel_ablation} shows the ablation study for dual-channel knowledge transfer. Belief transfer alone provides negligible improvement (+0.6\% reward, 3\% fewer episodes to reach Pure RL reward), indicating that spatial knowledge informs \textit{where} uncertainty exists but not \textit{how} to act. Buffer seeding alone yields substantial gains (+7.7\% reward, 33\% fewer episodes to reach Pure RL reward) by providing LGCP-demonstrated trajectories that accelerate policy learning. Notably, combining both channels achieves highest reward improvement and 38\% fewer episodes to reach Pure RL reward. This demonstrates that while belief transfer has minimal effect in isolation, it provides meaningful benefit when used in conjunction with replay buffer seeding. We interpret this as the spatial context helping the agent generalize beyond the specific trajectories in the buffer, enabling more effective exploration of high-value regions identified by the LGCP posterior.

\subsection{Warm start and Planning Horizon Ablations}
\label{Warm start and Planning Horizon Ablations}

\begin{figure}[!t]
    \centering
    \includegraphics[width=3.5in]{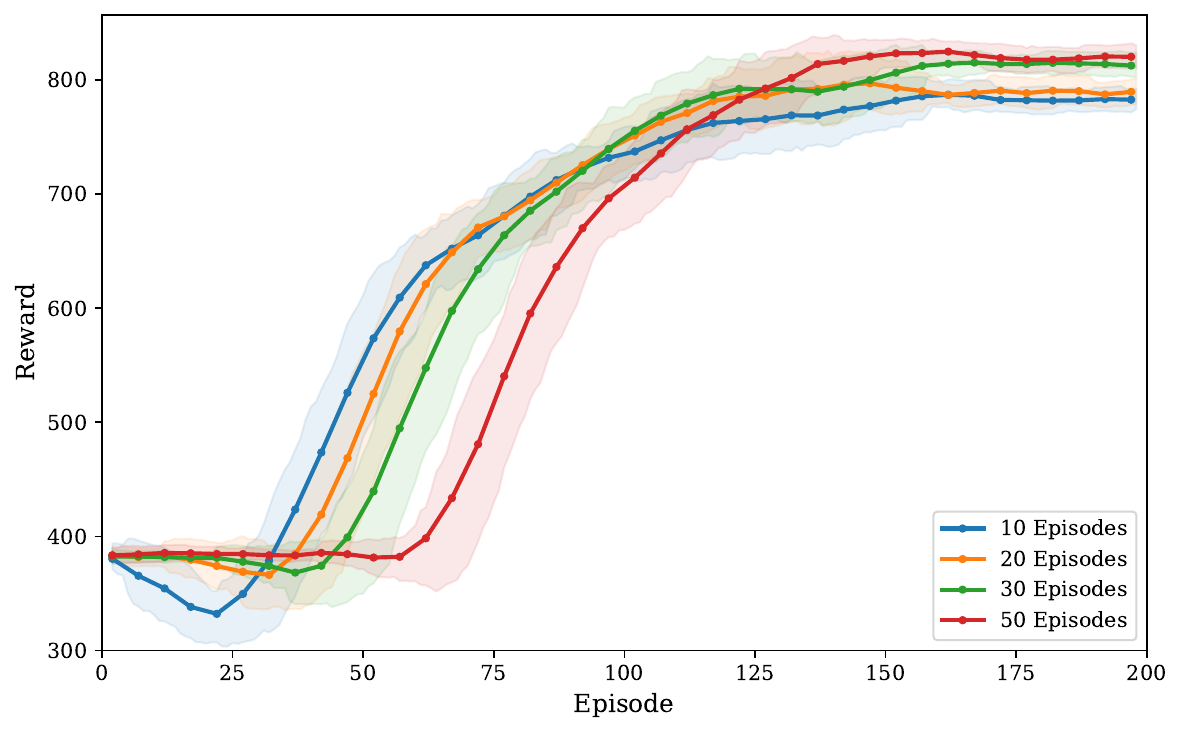}
    \caption{Effect of LGCP warm-start duration on SAC training. Different warm-start lengths determine the transition point from LGCP exploration (Phase-1) to SAC optimization(Phase-2).}\label{warm_start_ablation}
\end{figure}

The warm-start ablation in Figure~\ref{warm_start_ablation} evaluates how the duration of the LGCP exploration phase influences downstream SAC performance. All configurations maintain similar performance during their respective LGCP phases, with rewards stabilising around 380. The critical differences emerge during and after the policy transition. The 10-episode configuration exhibits a pronounced performance dip as it shifts to SAC, dropping approximately 12\% before recovering. This degradation occurs because the shorter warm-start phase collects only $2000$ replay buffer transitions, leaving the SAC agent with limited expert demonstrations to learn from. In contrast, longer warm-start durations accumulate more diverse state-action-reward transitions covering broader regions of the service area, enabling more effective policy initialisation.

A short warm-start of 10 episodes also yields the lowest final reward, as the replay buffer contains limited behavioural structure. Increasing the warm-start duration increases the final reward, with 20 and 30 episode warm-up yielding a clear improvement in rewards of approximately 6\% and 5.2\%, respectively. This occurs because longer LGCP exploration populates the replay buffer with a more diverse set of informative trajectories, enabling SAC to begin learning from substantially better initial conditions. However, the final reward of 30 episode vs 50 episode warm-start is identical. This is most likely attributable to excessive warm-starting which biases the replay buffer heavily towards LGCP-style behaviour, making SAC slower to adapt its own policy and resulting in a lagging training graph.

\begin{figure}[!t]
    \centering
    \includegraphics[width=3in]{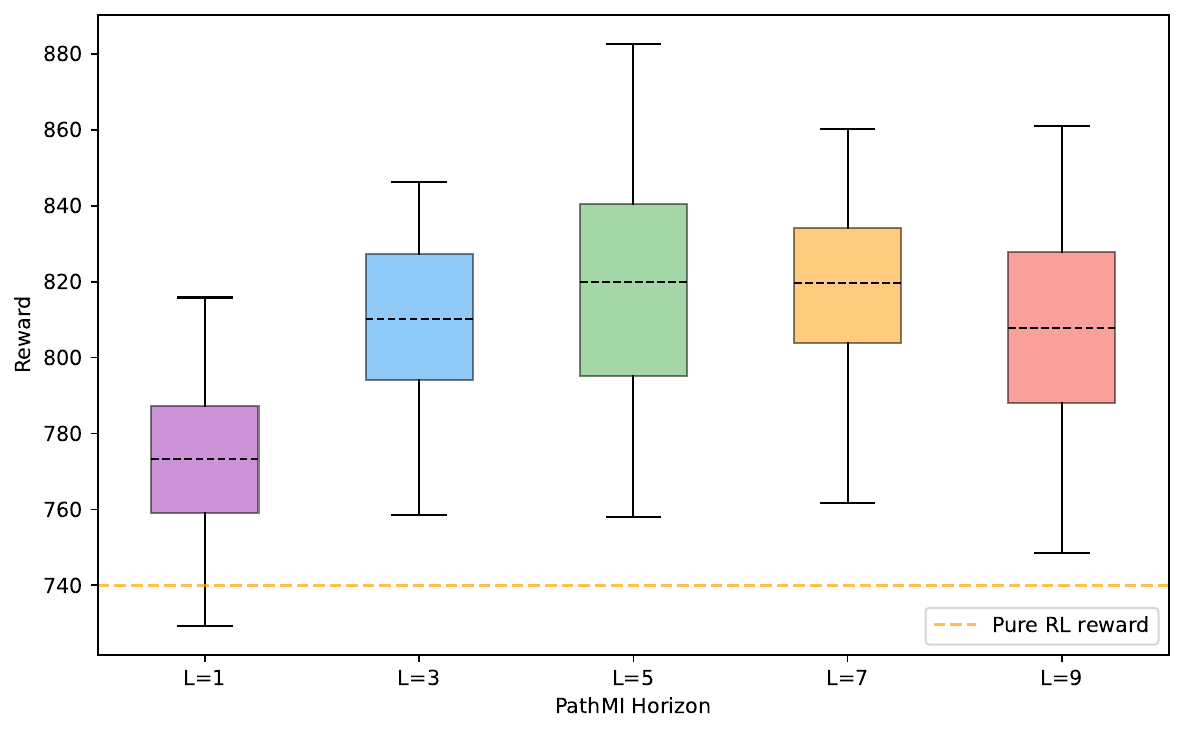}
    \caption{Effect of PathMI planning horizon on final reward.}\label{pathmi_horizon_scaled}
\end{figure}

The LGCP PathMI horizon ablation in Figure~\ref{pathmi_horizon_scaled} demonstrates that planning depth strongly influences final reward. The one-step myopic setting (L=1) achieves a substantially low median reward as compared to non-myopic configurations, as single-step planning greedily selects actions based on immediate information gain without considering future observational opportunities. Nevertheless, even this purely local strategy outperforms the pure RL baseline (740) by approximately 4.5\%. Increasing the horizon to L=3 and L=5 yields an improvement of 4.8\% and 6.1\%, respectively. These results indicate that moderate non-myopic planning enables UAVs to anticipate future information gain and commit to globally beneficial trajectories. However, extending beyond L=5 produces diminishing returns. L=7 maintains comparable performance while adding to the computational cost. L=9 degrades slightly, as uncertainty in long-horizon predictions reduces their reliability.

\subsection{Scalability and Coordination Analysis}
\label{Scalability and coordination analysis}

\begin{figure}[!t]
    \centering
    \includegraphics[width=3.5in]{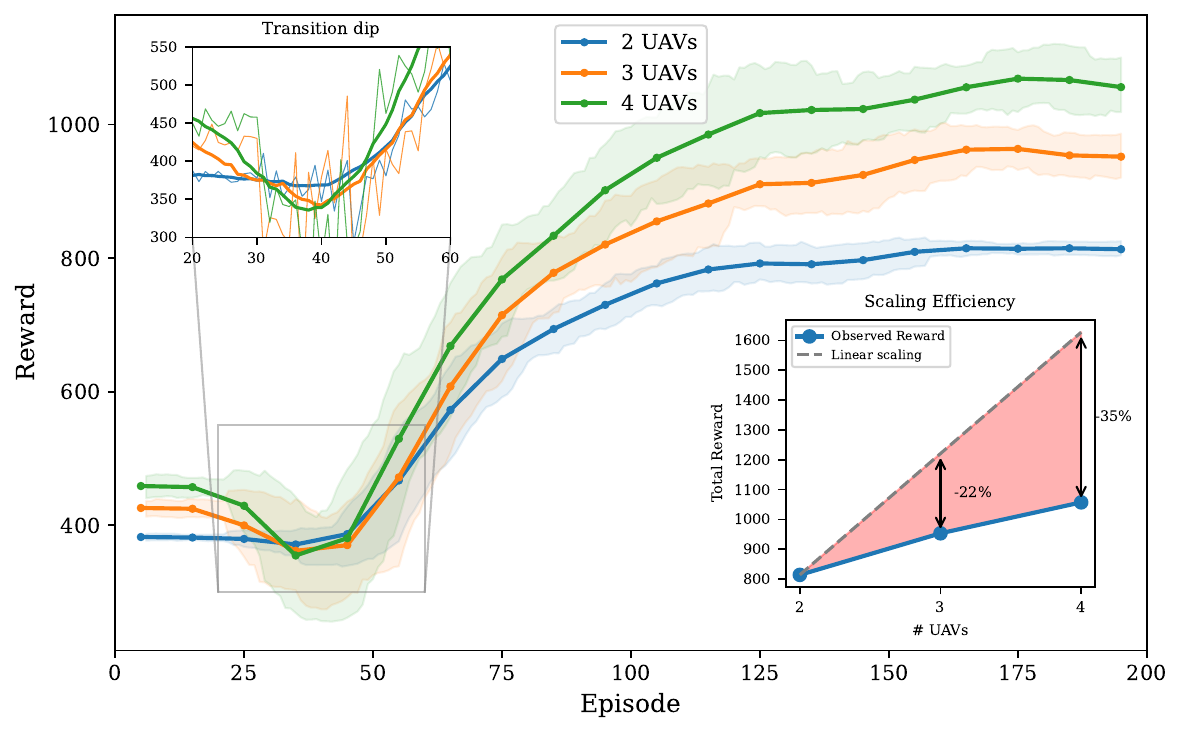}
    \caption{Learning performance comparison under varying numbers of UAVs. Increasing the number of agents improves overall reward but exhibits sub-linear scaling due to coordination overhead and redundant coverage}\label{uav_scaling}
\end{figure}

Figure~\ref{uav_scaling} evaluates the scalability of HBRL across multi-UAV configurations. The main plot demonstrates that all configurations successfully learn coordinated coverage policies, with final rewards of 820, 960, and 1080 for 2, 3, and 4 UAVs, respectively. The upper-left inset highlights the transition dip phenomenon occurring during episodes 30-50, where performance temporarily degrades as the SAC policy takes over from PathMI-guided exploration. It can be seen that this degradation is more pronounced for larger fleets, reflecting the increased complexity of learning coordinated behaviors when more agents must avoid spatial overlap. The lower-right inset quantifies scaling efficiency by comparing observed total reward against ideal linear scaling. The sublinear returns can be attributable to increasing coordination overhead, which is captured by the variance-normalized overlap penalty that discourages redundant coverage of already well-characterized regions. 

\begin{figure}[b]
    \centering
    \includegraphics[width=3.5in]{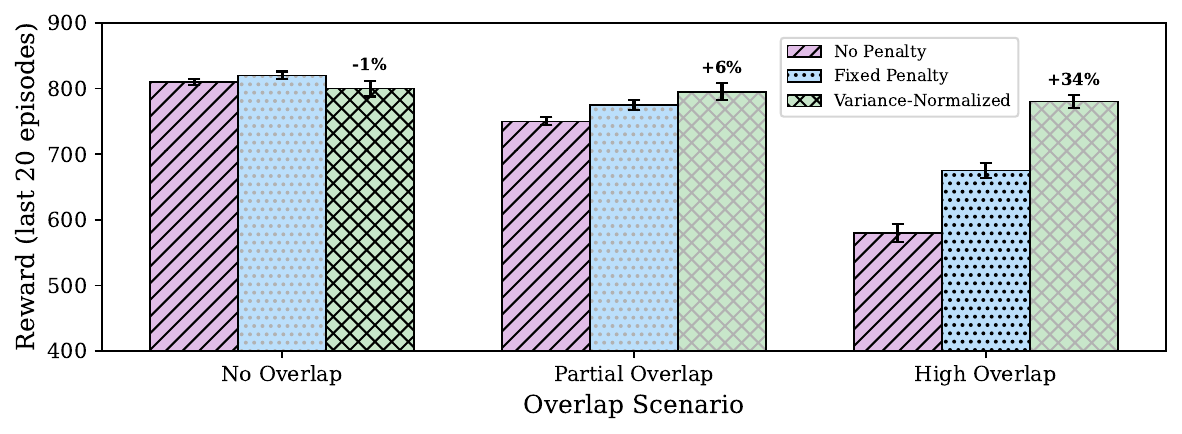}
    \caption{Comparison of reward under various overlap penalty scenarios.}\label{overlap_penalty_combined}
\end{figure}

Figure~\ref{overlap_penalty_combined} compares final performance under three levels of operational-region overlap and different coordination strategies. When the operational regions do not overlap, all methods achieve similar rewards, and both fixed and variance-normalized penalties have negligible impact as UAV sensing footprints rarely overlap. In partial overlap, coordination becomes beneficial: the fixed penalty yields improvements by discouraging some redundant sensing, while the proposed variance-normalized penalty performs best, improving the final reward by approximately 6\%. This gain arises from selectively allowing overlap in high-uncertainty regions while penalizing redundancy elsewhere. The benefit of adaptive coordination is most pronounced in the high-overlap regime, where substantial overlap between operational regions leads to frequent sensing conflicts and substantial performance degradation in the absence of coordination. In this case, the variance-normalized penalty improves final reward by about $34\%$ relative to no penalty, and by more than $14\%$ compared to the fixed penalty.

\subsection{Belief Dynamics and Uncertainty Calibration}
\label{Belief Dynamics and unvertainty Calibration}

\begin{figure}[!t]
    \centering
  \includegraphics[width=3.5in]{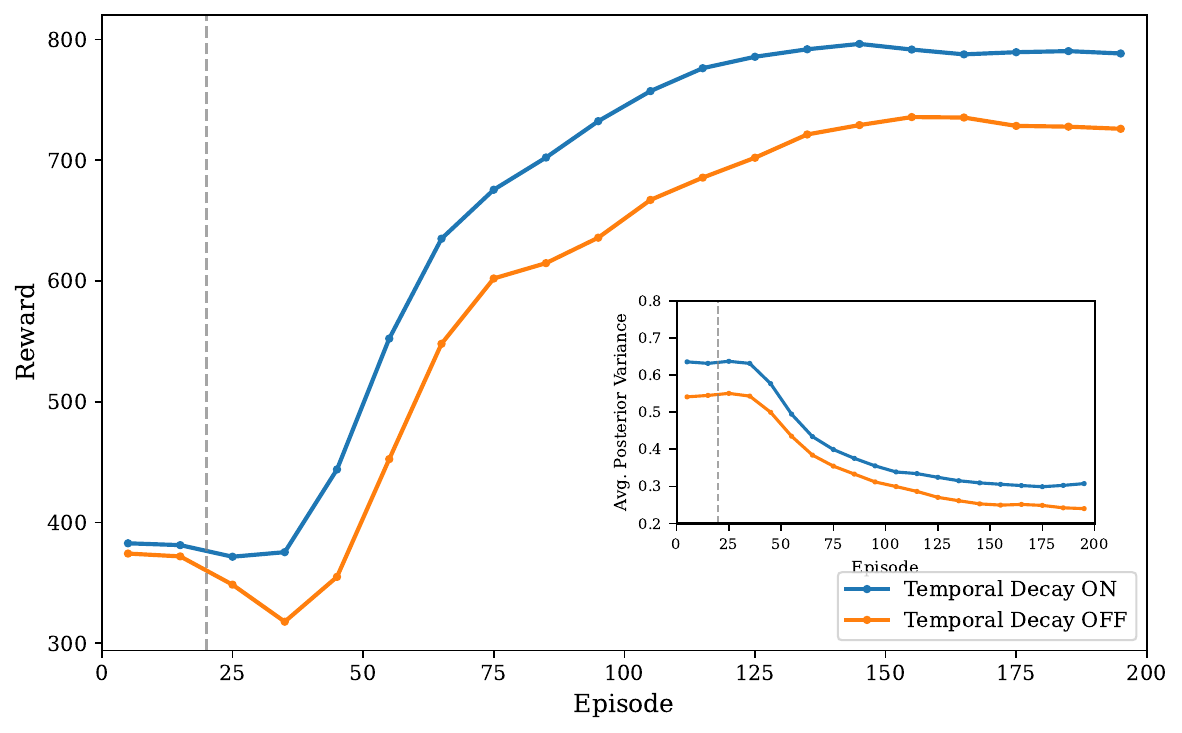}
  \caption{Impact of temporal decay on HBRL performance: (a) reward convergence and (b) belief uncertainty evolution. The dashed line denotes the warm-start transition point.}
  \label{temporal_decay_ablation}
\vspace{-4mm}
\end{figure}

Figure~\ref{temporal_decay_ablation} examines the role of temporal belief decay in maintaining accurate spatial awareness. Temporal decay increases posterior variance for cells not recently visited, prompting revisiting previously visited areas. Enabling temporal decay achieves 8.2\% higher final reward, demonstrating the value of maintaining fresh beliefs. However, it reveals a key insight: with temporal decay turned on, the system reports higher posterior variance because it correctly acknowledges uncertainty growth in unvisited regions. This elevated variance drives the planner to revisit areas with stale information, keeping beliefs calibrated with updated demand patterns. Conversely, with no temporal decay of information, variance remains artificially low as the system falsely assumes past observations remain valid indefinitely. This leads to no revisitation despite information staleness, resulting in decisions based on outdated beliefs and, consequently, lower cumulative reward.

\begin{figure*}[!t]
    \centering
    \includegraphics[width=5in]{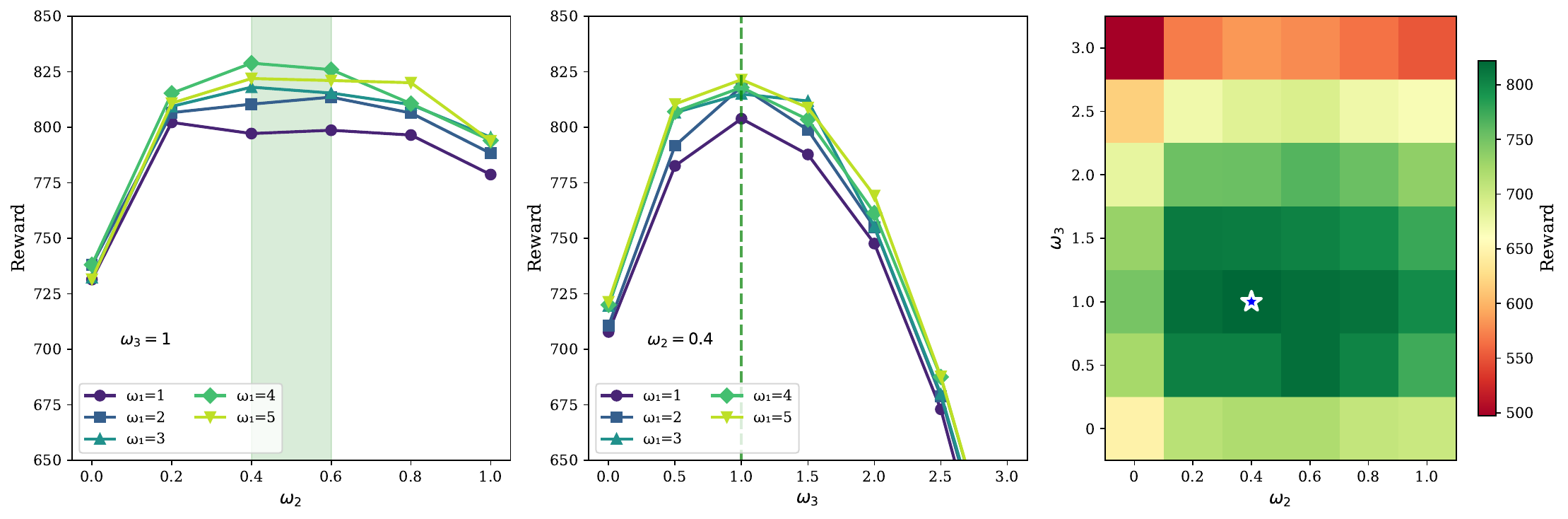}
    \caption{Reward weight sensitivity analysis. (a) Effect of exploration weight $\omega_2$ on final reward with coordination weight fixed at $\omega_3 = 1.0$. The shaded region indicates the optimal range $\omega_2 \in [0.4, 0.6]$. (b) Effect of coordination weight $\omega_3$ on final reward with exploration weight fixed at $\omega_2 = 0.4$. (c) Reward heatmap over the $(\omega_2, \omega_3)$ configuration space with $\omega_1 = 5$. The star indicates the default configuration.}\label{weight_sensitivity_combined}
\end{figure*}

Figure~\ref{weight_sensitivity_combined} presents the sensitivity analysis for reward weights $(\omega_1, \omega_2, \omega_3)$. The exploration weight $\omega_2$ exhibits a broad optimal plateau between 0.4 and 0.6, with $\omega_2 = 0$ causing 11\% degradation due to insufficient exploration incentive. Similarly, while extreme $\omega_3$ values degrade performance, the optimal region spans a broad range, which indicates robustness to moderate tuning variations. The service weight $\omega_1$ shows best and comparable reward for $\omega_1 = 4,5$, with $\omega_1 = 6$ performing worse, indicating that over-emphasis on immediate service rewards causes the agent to exploit known hotspots at the expense of discovering new demand regions. 

\subsection{Robustness to Experience Loss}
\label{robustness to observation loss}

\begin{figure}
    \centering
    \includegraphics[width=3.5in]{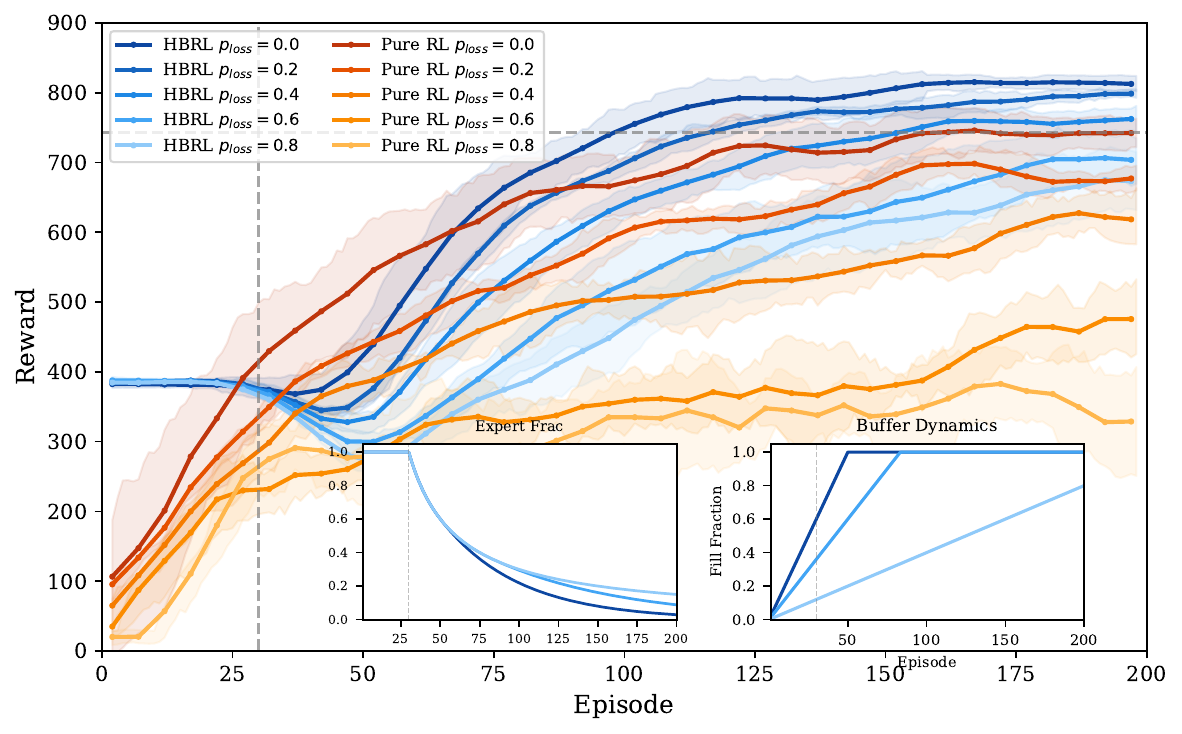}
    \caption{Training curves for different $p_{\mathrm{loss}}$ values}\label{experience_loss_learning_curves}
\end{figure}

\begin{figure}[b]
    \centering
    \includegraphics[width=\columnwidth]{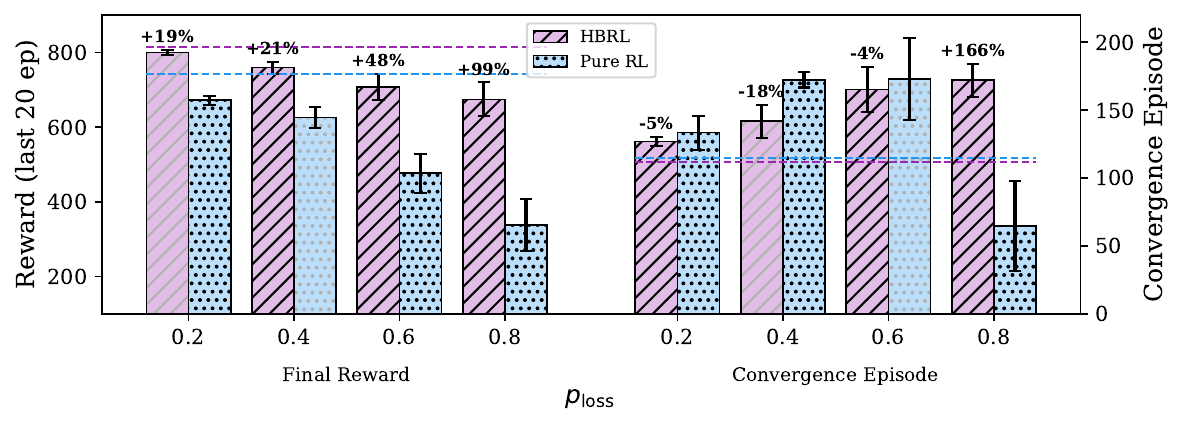}
    \caption{Final Reward and Convergence vs. $p_{\mathrm{loss}}$}\label{experience_loss_bar_charts}
\end{figure}

Figure~\ref{experience_loss_learning_curves} evaluates robustness under stochastic replay corruption. We replace deterministic replay insertion with independent Bernoulli thinning, where each transition $(\mathbf{s}_t,\mathbf{a}_t,r_t,\mathbf{s}_{t+1})$ is inserted into the buffer with probability $1-p_{\mathrm{loss}}$, for $p_{\mathrm{loss}} \in \{0.0,0.2,0.4,0.6,0.8\}$. All results are averaged over five random seeds, paired across all methods and $p_{\mathrm{loss}}$ values. Performance degrades smoothly as $p_{\mathrm{loss}}$ increases, indicating graceful degradation under reduced replay availability. At moderate loss levels ($p_{\mathrm{loss}}\le 0.4$), both methods maintain stable learning dynamics, while higher loss primarily slows convergence due to delayed buffer saturation. The inset plots show that increasing $p_{\mathrm{loss}}$ reduces effective buffer fill-rate, thereby limiting the diversity of stored transitions. Figure~\ref{experience_loss_bar_charts} summarizes final performance over the last 20 episodes. Across all loss levels, degradation is gradual rather than catastrophic, demonstrating the robustness of the proposed framework to experience/replay loss. Convergence here is defined as the first episode reaching $95\%$ of each configuration's \textit{own} final reward.

\FloatBarrier
\section{Conclusion}
This paper presents HBRL, a belief-guided warm-start reinforcement learning framework for coordinating multiple mobile agents that must act on incomplete and evolving knowledge of a spatial demand field. The framework integrates LGCP-based spatial inference with non-myopic information-driven planning and off-policy deep reinforcement learning. The approach enables coordinated policy learning while accounting for uncertainty in spatial demand. Performance was evaluated by comparison with planning-only and learning-only baselines across different overlap configurations, workload intensities, and training settings. The results show that the HBRL reaches the Pure SAC reward level in fewer episodes and achieves consistently higher long-term reward, with substantial reward and convergence speed improvements compared to baseline approaches. The framework is well-suited to multiple mobile agent settings where shared spatial structure can be exploited to accelerate learning, as demonstrated at fleet sizes up to $N{=}4$ in the wireless setting (Section~\ref{Scalability and coordination analysis}) and up to $N{=}10$ on a second, real-data domain (Appendix~\ref{Appendix_C}).

However, this work is restricted to the initial warm-start phase, which suggests extending the framework toward continuous belief--policy co-adaptation as a direction for future work. Additionally, the impact of scaling to significantly larger fleet sizes merits further investigation. The simulated reward is a proxy composed of service, exploration, and coordination terms weighted by $(\omega_1, \omega_2, \omega_3)$. Its correspondence to application-specific deployment utility is a modelling assumption that warrants validation in relevant deployment scenarios. Sim-to-real transferability and the bridging the gap between simulated policies and behaviour on real hardware remains an open direction for this framework, where hardware-in-the-loop evaluation may be a natural next step.

Beyond the two instantiations evaluated here, the core framework components are domain-agnostic: LGCP posterior inference, PathMI planning, dual-channel knowledge transfer, and the variance-normalized overlap penalty may apply to a broad range of spatially heterogeneous multi-agent exploration settings. Only the observation model and sensing-footprint geometry are task-specific and require adaptation when moving to a new domain.

\section{Acknowledgment}
This work was supported in part by the Commonwealth Scholarship Commission, U.K., and in part by the Communications Hub for Empowering Distributed ClouD Computing Applications and Research (CHEDDAR) funded by the UKRI Engineering and Physical Sciences Research Council (EPSRC) under Grant EP/X040518/1. The authors acknowledge the use of generative AI tools (OpenAI ChatGPT and Grammarly) to assist with language refinement. All conceptual development, experimental design, analysis, and conclusions were conducted and verified by the authors.

{\footnotesize
\bibliographystyle{unsrt}
\bibliography{main}}
\clearpage
\appendices
\FloatBarrier
\section{Warm Start Exploration Strategies}
\label{Appendix_A}
To motivate the choice of PathMI planning for Phase~1, we compare four action-selection methods in a planning-only setting (without RL). All strategies share the same belief model and differ only in trajectory planning: \textbf{PathMI} (multi-step lookahead via Equation~\ref{eq:pathmi_score}), \textbf{UCB} (greedy, $\mu_c + \kappa\sqrt{\sigma^2_c}$, $\kappa{=}2$), \textbf{Lawnmower} (boustrophedon sweep), and \textbf{Random} (uniform direction). To isolate the planning strategy from other framework components, we evaluate on a $1000 \times 1000$\,m scenario ($\Delta{=}10$\,m, two Gaussian hotspots, $N{=}2$ UAVs, $r_c{=}100$\,m and 10 seeds per configuration).

\begin{figure*}[!t]
    \centering
    \begin{minipage}[t]{0.48\textwidth}
        \centering
        \includegraphics[width=3in]{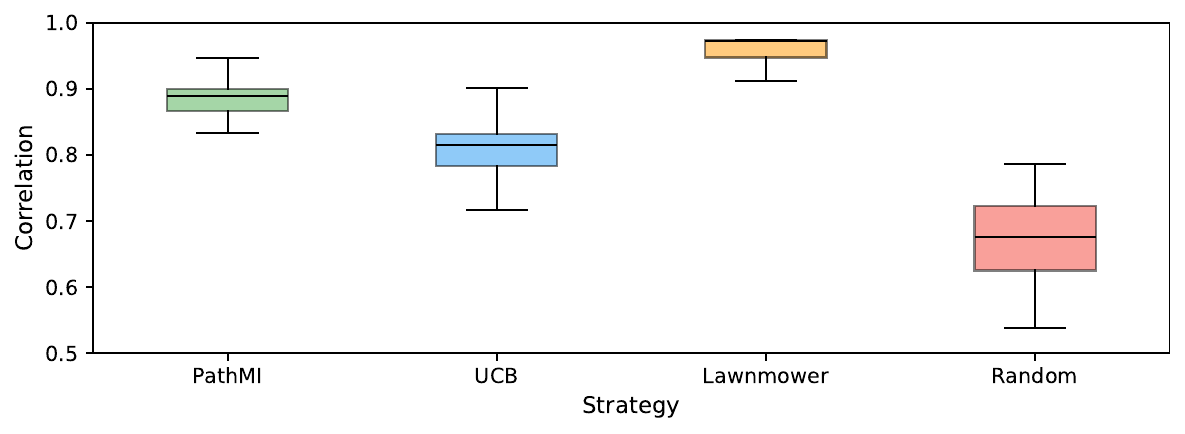}
        \captionof{figure}{Strategy vs. Correlation}\label{strategy_correlation}
    \end{minipage}\hfill
    \begin{minipage}[t]{0.48\textwidth}
        \centering
        \includegraphics[width=3in]{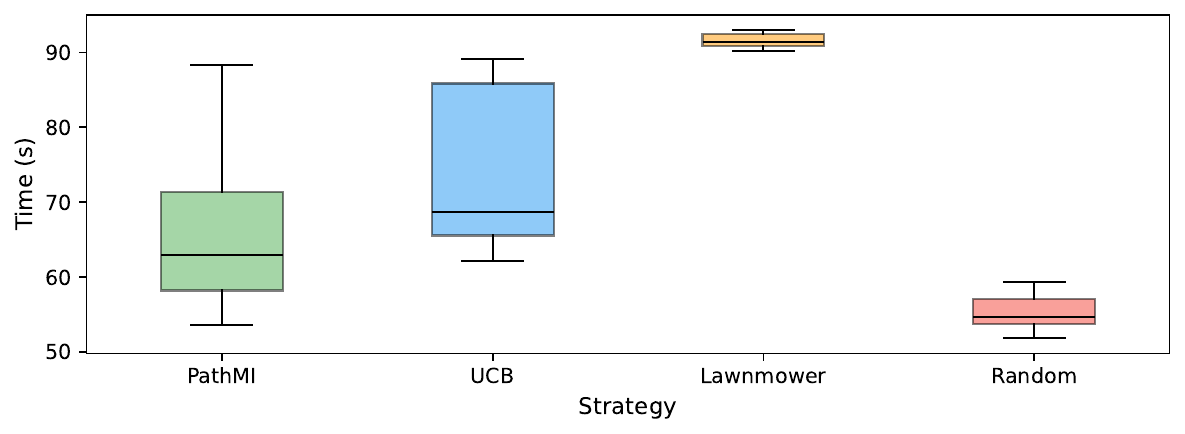}
        \captionof{figure}{Strategy vs. Runtime}\label{strategy_runtime}
    \end{minipage}
\end{figure*}
\begin{figure*}[!b]
    \centering
    \includegraphics[width=6.5in]{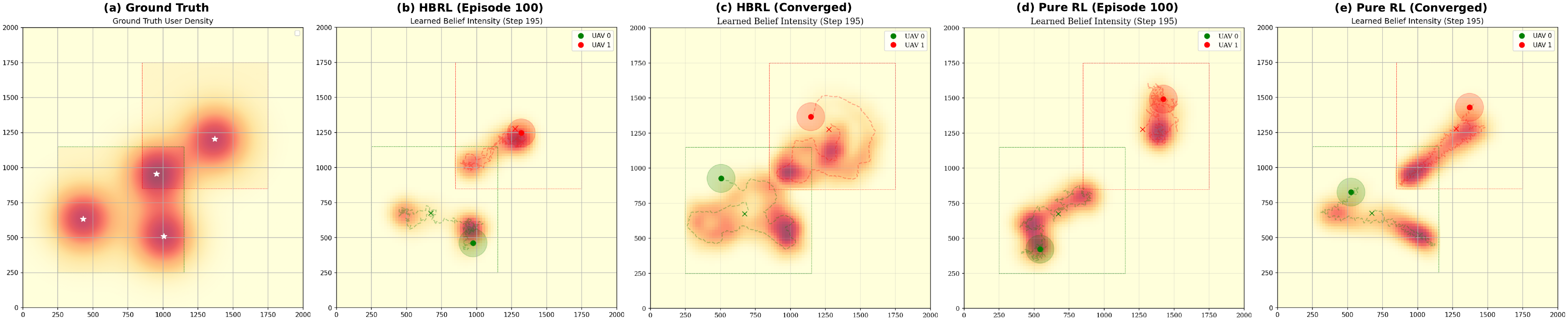}
    \caption{Learned belief intensity maps at episode 100 and convergence for HBRL and Pure RL, compared against
   the ground truth}\label{gt_vs_belief_comparison}
\end{figure*}
Figures~\ref{strategy_correlation} and~\ref{strategy_runtime} report Pearson correlation between estimated and ground-truth intensity and per-episode runtime. PathMI achieves median correlation $0.89$, approaching Lawnmower ($0.96$) while requiring around $30\%$ less time. UCB attains $0.81$ with higher variance due to greedy local optima. Lawnmower achieves the highest correlation but cannot adapt to observed demand. These results support PathMI as the Phase~1 strategy, offering the best accuracy--computation trade-off among adaptive methods.

\section{Learned Belief Intensity}
\label{Appendix_B}
Figure~\ref{gt_vs_belief_comparison} presents the final belief intensity maps reconstructed by each method at selected training episodes. At episode 100, HBRL has already identified all four density centers across both operational regions, producing a belief map closely resembling the ground truth. In contrast, Pure RL at episode 100 has only discovered one density center per region, reflecting its slower learning and confirming that the performance gap is due to sample efficiency. Animated visualizations of the belief evolution over the course of each episode are provided as supplementary material.

\section{Agreement Between the Diagonal Proxy and the Full Marginal Variance}
\label{Appendix_D}

Section~\ref{sec:laplace} adopts the diagonal Laplace proxy $\sigma_c^2 = 1/[\mathbf{H}]_{cc}$ (Equation~\ref{eq:diagonal_variance}) in place of the exact marginal variance $[\mathbf{H}^{-1}]_{cc}$, whose computation requires sparse matrix inversion that is prohibitive at operational grid sizes. Because exploration decisions depend only on the relative ordering of cell uncertainties, the relevant question is whether the proxy preserves this ordering. We verify this empirically on grids small enough for direct inversion, using the main experimental parameters ($\tau = 1.0$, $\beta = 0.2$; Table~\ref{tab:parameters}).

\begin{table*}[!t]
\begin{center}
\begin{tabular}{lrcccc}
\textbf{Grid} & $|\mathcal{G}|$ & \textbf{Spearman} $\rho$ & \textbf{PLA} ($L{=}3$) & \textbf{PLA} ($L{=}5$) & $t_{\text{full}}/t_{\text{diag}}$ \\
\hline \\[-0.9em]
$5\times 5$ & $25$ & $0.985 \pm 0.009$ & $0.990$ & $0.990$ & $\sim\!60\times$ \\
$10\times 10$ & $100$ & $0.979 \pm 0.008$ & $0.990$ & $0.985$ & $\sim\!90\times$ \\
$15\times 15$ & $225$ & $0.971 \pm 0.007$ & $0.985$ & $0.975$ & $\sim\!290\times$ \\
$20\times 20$ & $400$ & $0.964 \pm 0.006$ & $0.993$ & $0.993$ & $\sim\!510\times$ \\
$30\times 30$ & $900$ & $0.957 \pm 0.004$ & $0.990$ & $0.990$ & $\sim\!1300\times$ \\
\end{tabular}
\end{center}
\caption{Agreement between the diagonal proxy and the full marginal variance (200 scenarios per row for grids up to $15\times 15$, 150 for $20\times 20$, 100 for $30\times 30$).}
\label{tab:proxy_agreement}
\end{table*}

Table~\ref{tab:proxy_agreement} reports two quantities. Spearman's rank correlation $\rho$ measures cell-level agreement between the two variance vectors. Path-level agreement (PLA) is the fraction of scenarios in which $\arg\max$ PathMI (Equation~\ref{eq:best_path}) selects the same direction under both variance estimates, at horizons $L = 3$ and $L = 5$. The time ratio $t_{\text{full}}/t_{\text{diag}}$ is the per-call wall-clock cost of direct inversion relative to the proxy. Spearman $\rho \geq 0.95$ across all grids demonstrates strong cell-level rank agreement, and PLA $\geq 0.975$ at both horizons shows that the exploration action selected by PathMI is identical under both estimates in at least $97.5\%$ of cases. The cost ratio grows from $\sim\!60\times$ at $5\times 5$ to $\sim\!1300\times$ at $30\times 30$, supporting the use of the proxy at the operational grid size ($100 \times 100$).

\FloatBarrier
\section{From Expected Information Gain to PathMI}
\label{Appendix_E}

The information-optimal path maximizes the expected information gain of the observations collected along the path~\citep{rainforth2024modern},
\begin{equation}
    \mathcal{P}^{\star} = \arg\max_{\mathcal{P}} \; I\left(\mathbf{u};\, \mathbf{y}_{\mathcal{O}(\mathcal{P})} \mid \mathcal{D}_t\right),
    \label{eq:exact_ig}
\end{equation}
where $\mathcal{O}(\mathcal{P})$ is the set of cells observed along $\mathcal{P}$ (Equation~\ref{eq:path_coverage}) and $\mathcal{D}_t$ denotes the data observed up to time $t$. For the Poisson--log-Gaussian pair, exact evaluation requires marginalization over future observations and is intractable for online planning over many candidate paths. PathMI follows from Equation~\ref{eq:exact_ig} under the following approximations.

\textbf{(i) Laplace--Gaussian posterior with plug-in information.} Under the Laplace approximation of Section~\ref{sec:laplace}, the posterior over $\mathbf{u}$ is Gaussian with covariance $\boldsymbol{\Sigma}_u = \mathbf{H}^{-1}$. For $y_c \sim \text{Poisson}(\exp(u_c))$ the log-likelihood curvature is $-\exp(u_c)$ independently of $y_c$, so one observation contributes Fisher information $J_c = \lambda_c$ exactly, evaluated at the current posterior mode, the likelihood term of Equation~\ref{eq:hessian}. With this plug-in, the expected entropy reduction of the Gaussian belief has the closed form~\citep{MacKay1992ActiveData}
\begin{equation}
    I\left(\mathbf{u};\, \mathbf{y}_{\mathcal{O}} \mid \mathcal{D}_t\right) \approx \frac{1}{2} \log\det\left(\mathbf{I} + \boldsymbol{\Sigma}_u \mathbf{J}\right), \qquad \mathbf{J} = \mathrm{diag}(J_c).
    \label{eq:gauss_ig}
\end{equation}

\textbf{(ii) Diagonal variance proxy.} Substituting the proxy $\mathrm{diag}(\sigma_c^2)$ of Equation~\ref{eq:diagonal_variance}, which drops the off-diagonal posterior couplings and replaces the marginal variance $[\mathbf{H}^{-1}]_{cc}$ with $1/[\mathbf{H}]_{cc}$, factorizes Equation~\ref{eq:gauss_ig} into the per-cell sum $\frac{1}{2}\sum_{c \in \mathcal{O}(\mathcal{P})} \log(1 + \lambda_c \sigma_c^2)$. Appendix~\ref{Appendix_D} verifies empirically that the variance-proxy component of this substitution preserves the planner's action ranking.

\textbf{(iii) Linearisation.} Linearising $\log(1+x) \approx x$ gives the score $\frac{1}{2}\sum_{c} \lambda_c \sigma_c^2$; the linearisation is monotone in each cell's contribution, and the constant $\frac{1}{2}$ is immaterial for ranking.

\textbf{(iv) Predictive-variance surrogate.} Following the predictive-variance criterion for active data selection ~\citep{MacKay1992ActiveData, Riis2022BayesianActive}, we drop the intensity weight $\lambda_c$, treating per-visit information as approximately uniform across candidate cells. This decouples the score from the current mode estimate, whose error is largest in unexplored regions. For observed cells $[\mathbf{H}]_{cc} = \lambda_c + Q_{cc}$, so $\sigma_c^2$ already carries an inverse dependence on $\lambda_c$ that partially offsets the dropped factor.

\textbf{(v) Repeat-visit discount.} The exposure-indicator Hessian of Equation~\ref{eq:hessian} accrues likelihood curvature from each observed cell's most recent observation only, and the variance update of Equation~\ref{eq:variance_update} overwrites rather than accumulates, so the stored variance saturates after a visit. The discount $\xi_c = 1/(1+n_c)$ of Equation~\ref{eq:xi_definition} restores the diminishing returns of repeated observation: holding the intensity estimate fixed, $n_c$ observations accumulate precision $n_c \lambda_c$ on the pre-observation precision $h_c$, so the linearised gain of the next observation is
\begin{equation}
    \Delta_c(n_c) = \frac{1}{2} \cdot \frac{\lambda_c}{h_c + n_c \lambda_c}
    = \frac{1}{2} \cdot \frac{\lambda_c \sigma_{c,0}^2}{1 + n_c\, \lambda_c \sigma_{c,0}^2}, \qquad \sigma_{c,0}^2 = 1/h_c.
    \label{eq:marginal_gain}
\end{equation}
$\xi_c$ matches the discount factor of Equation~\ref{eq:marginal_gain} under the unit-information normalization $\lambda_c \sigma_{c,0}^2 = 1$ and reproduces its asymptotic $1/n_c$ decay in general.

\textbf{Staleness bonus.} The additive term $\omega_s \min(s_c/s_{\max}, 1)\, \sigma_c^2$ is a revisit incentive complementing the variance dynamics of Section~\ref{sec:temporal}: variance inflation under Equation~\ref{eq:variance_predict} is gradual and capped, and the bonus adds an explicit, bounded incentive to revisit long-unobserved cells within the planning horizon; $\omega_s$ is a tuned weight.

\begin{table}[!t]
\begin{center}
\begin{tabular}{lccc}
\textbf{Grid} & \textbf{Same direction} & \textbf{Retained value} & \textbf{95th-pct.\ loss} \\
\hline \\[-0.9em]
$20\times20$, $L{=}3$ & $0.74 \pm 0.01$ & $0.991$ & $0.050$ \\
$20\times20$, $L{=}5$ & $0.71 \pm 0.01$ & $0.990$ & $0.053$ \\
$30\times30$, $L{=}3$ & $0.73 \pm 0.02$ & $0.993$ & $0.035$ \\
$30\times30$, $L{=}5$ & $0.73 \pm 0.02$ & $0.992$ & $0.040$ \\
\end{tabular}
\end{center}
\caption{PathMI versus the plug-in information gain of Equation~\ref{eq:gauss_ig}, averaged over observed fractions $f\in[0,0.30]$ and five seeds. Same direction: fraction of states in which both select the same candidate (mean $\pm$ std). Retained value: Equation~\ref{eq:gauss_ig} value of the PathMI-selected path relative to the best candidate. 95th-pct.\ loss: 95th percentile of one minus the retained value.}
\label{tab:pathmi_eig}
\end{table}
To quantify the effect of approximations (ii)--(v) on decisions, we compared the direction selected by PathMI with that of the plug-in information gain of Equation~\ref{eq:gauss_ig}, evaluated with the full posterior covariance, on $20\times20$ and $30\times30$ grids with the parameters used in Table~\ref{tab:proxy_agreement}, over belief states with observed fractions from $0$ to $0.30$ (Table~\ref{tab:pathmi_eig}). Equation~\ref{eq:gauss_ig} is itself a faithful reference: a Monte Carlo estimate of the exact gain of Equation~\ref{eq:exact_ig} selects the same direction as Equation~\ref{eq:gauss_ig} in at least $90\%$ of cases. The two increasingly select different directions as observations accumulate, but the alternatives are nearly tied: the PathMI choice retains at least $0.99$ of the Equation~\ref{eq:gauss_ig} value on average at every observed fraction. The diagonal step (ii) changes the selected direction in under one percent of cases, consistent with Appendix~\ref{Appendix_D}, and the dropped intensity weight (iv) accounts for most of the difference.

\FloatBarrier
\section{A Second Instantiation: Mobility-on-Demand Sensing}
\label{Appendix_C}

To demonstrate that the framework is not specific to the aerial-wireless setting, we instantiate it on a structurally and operationally distinct domain: a fleet of ground vehicles acting as mobile sensors that reconstruct a city-scale demand field from real New York City taxi records. This experiment evaluates transfer to a distinct domain using measured real-world demand data. The belief model, PathMI planner, dual-channel transfer, action space, state space, and reward are reused without modification; an agent is now an idle vehicle rather than a UAV, the demand field is measured rather than generated, and the spatial extent is nearly an order of magnitude larger. Only the demand source and a small set of scale constants differ.

\subsection{Environment}

\begin{figure*}[!t]
    \centering
    \includegraphics[width=4in]{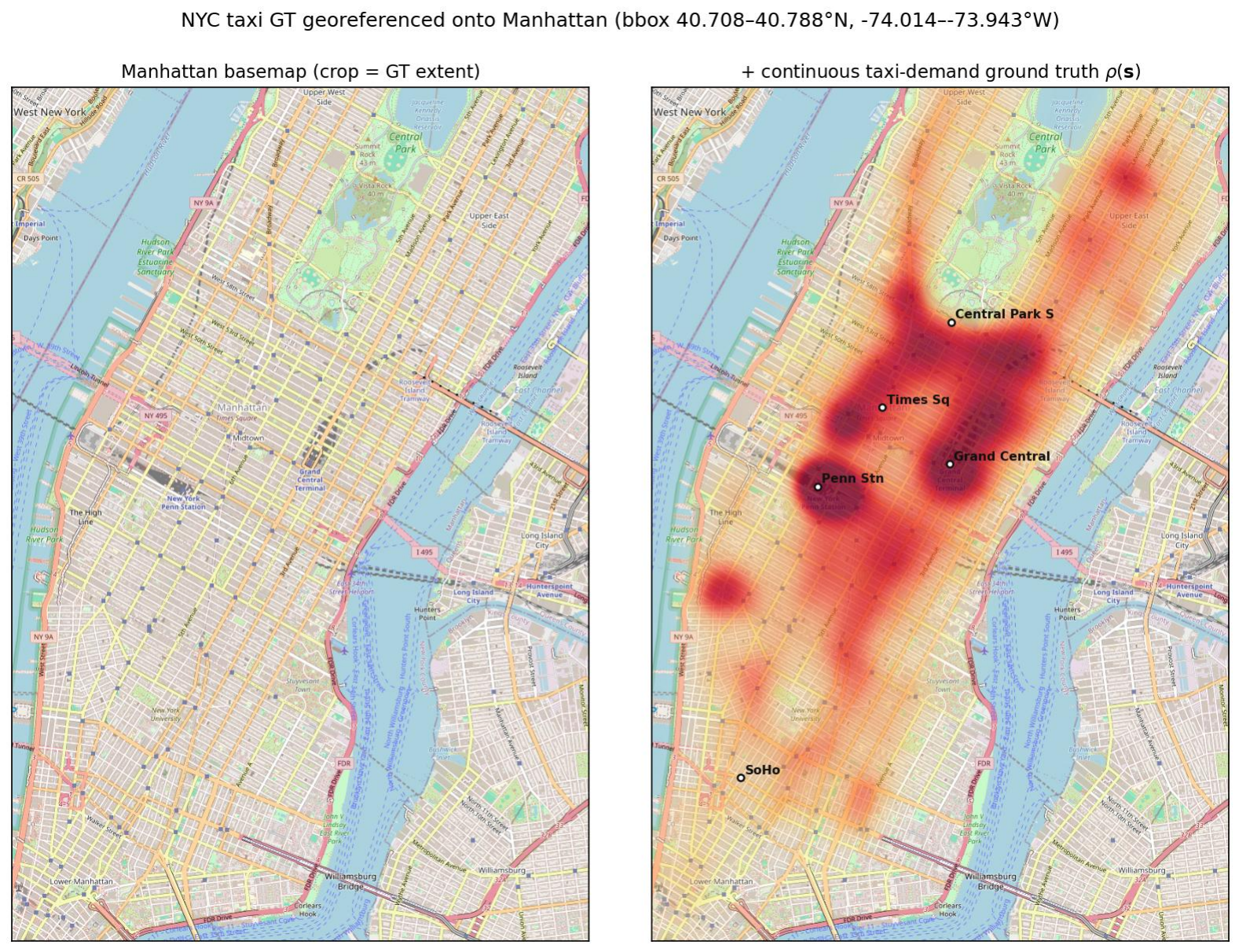}
    \caption{The mobility scene. Left: the Manhattan operational district (the crop defines the ground-truth extent). Right: the continuous taxi-demand field $\rho(\mathbf{s})$ recovered from real NYC pickups, georeferenced onto the map. Demand concentrates along the Midtown--downtown corridor (Times~Square, Penn~Station, Grand~Central) and reduces near the rivers, consistent with known Manhattan demand hotspots.}
    \label{fig:mobility_overlay}
\end{figure*}
We consider a Manhattan operational district of $8.9 \times 6.0$\,km (Figure~\ref{fig:mobility_overlay}). The rider demand $\rho(\mathbf{s})$ (Equation~\ref{eq:density_field}) is the latent intensity of a spatial point process, which we recover by kernel-smoothing publicly available NYC~TLC yellow-taxi pickups~\citep{NYC_TLC} onto a $178 \times 120$ grid of $50$\,m cells, with water cells masked out of the operational area. Observations are generated exactly as in the wireless instantiation: observed counts follow the binomial model of Equation~\ref{eq:observation_model} with unchanged $n_s$ and $\pi_{\text{req}}$, with the recovered demand field, normalized to $[0,1]$ via $\rho_{\max}$, taking the place of the Gaussian mixture of Equation~\ref{eq:density_field}, so the demand field is measured rather than generated while the sensing process is identical across the two instantiations. Because the measured demand field is derived from real observations rather than the synthetic Gaussian-mixture generator used in Section \ref{System Model}, this instantiation evaluates whether HBRL generalizes beyond the synthetic demand-generation assumptions while learning demand online from observations. A vehicle observes request counts only within its sensing footprint $\mathcal{C}_i(t)$ (Equation~\ref{eq:coverage_footprint}), captured by the exposure variable $e_c(t)$ Equation~\ref{eq:exposure_indicator}: cells the fleet never visits remain unobserved, so it must move to learn the demand field. The action, state, and reward are inherited without change: the action $\mathbf{a}_i(t) \in [-1,1]^2$ maps to displacement through $\mathbf{v}_i = \mathbf{a}_i\, d_{\max}$ (Equation~\ref{eq:action_mapping}), the state $\mathbf{s}_t \in \mathbb{R}^{3N+4}$ (Equation~\ref{eq:state_space}) concatenates the normalized fleet pose with the shared-belief summary, and the reward retains the three-objective form of Equation~\ref{eq:reward}, in which $R_{\text{service}}$ now counts rider requests the fleet is positioned to sense, $R_{\text{explore}}$ rewards uncertainty reduction, and $C_{\text{coord}}$ penalizes redundant coverage. Vehicle motion follows the point-mass dynamics of Equation~\ref{eq:position_update}; routing on the street network is an application-level refinement outside the scope of the framework-level contribution.

\begin{table*}[!t]
\caption{Key settings changed for the mobility instantiation. All algorithmic components (belief, planner, SAC, dual-channel transfer, reward structure) are unchanged.}
\label{tab:mobility_changes}
\begin{center}
\begin{tabular}{lcc}
\textbf{Setting} & \textbf{UAV wireless} & \textbf{MoD (NYC taxi)} \\
\hline \\[-0.9em]
Agent & UAV & idle vehicle (mobile sensor) \\
Operational area & $2 \times 2$~km & $8.9 \times 6.0$~km (Manhattan) \\
Grid resolution $\Delta$ & $20$~m & $50$~m \\
Sensing radius $r_c$ & $100$~m & $200$~m \\
Max.\ displacement $d_{\max}$ & $15$~m/step & $432$~m/step (~29km/h) \\
Demand source & Gaussian mixture with drift (Equation~\ref{eq:density_field}) & measured NYC taxi, diurnal \\
Episode & $200$ steps & $200$ steps (54 s) over a $3$~h window \\
Warm-start episodes $K_w$ & $10$--$50$ & $20$ \\
\end{tabular}
\end{center}
\end{table*}
Only the demand model and a handful of scale constants are re-instantiated, as summarized in Table~\ref{tab:mobility_changes}. An episode spans a fixed $3$\,h window of a single day (we report the $23{:}00$--$02{:}00$ night window, over which the relative intensity of the hotspots shifts from the Midtown cluster toward downtown) discretized into $200$ control steps, over which the demand field advances continuously by interpolating between the measured hourly frames. The belief persists within an episode and is reset to the prior between episodes, and training and evaluation use disjoint day-windows. Consecutive hourly fields have median spatial correlation $0.95$ and their centre of mass moves only $120$\,m per hour, so the field is stable in shape while evolving in intensity, a slow within-episode drift similar to the wireless regime but driven here by a real daily demand cycle.

\subsection{Results}

\begin{figure}[!t]
    \centering
    \includegraphics[width=3.5in]{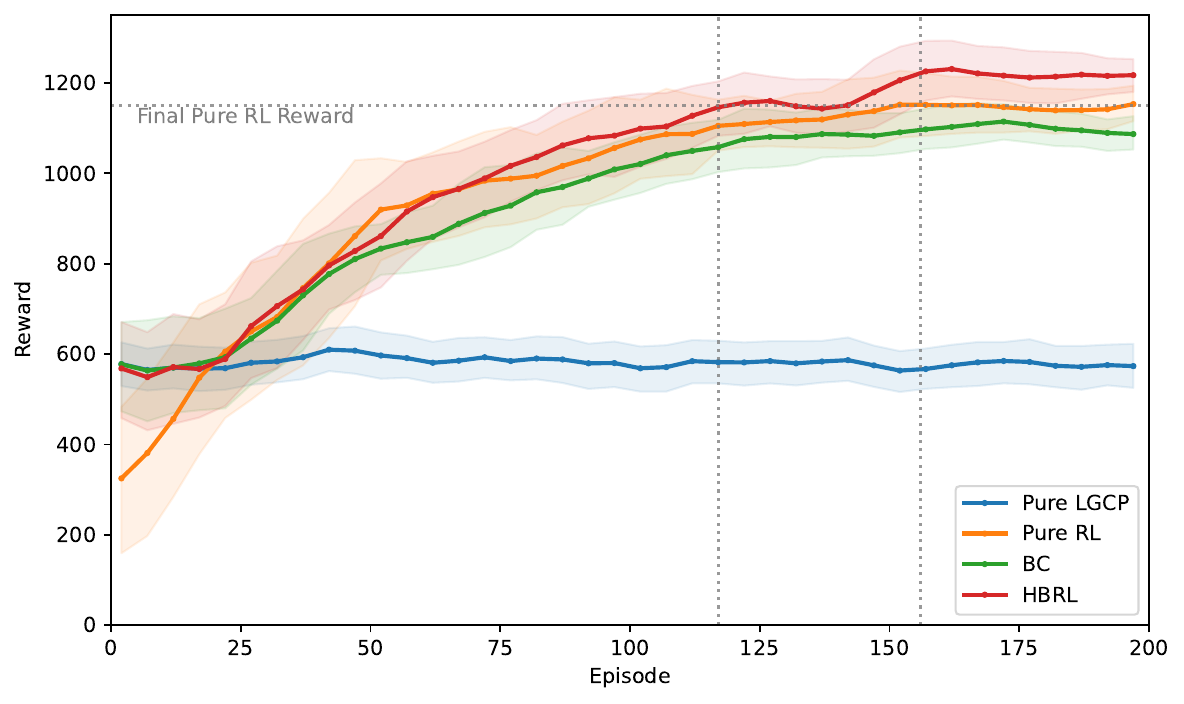}
        \caption{Reward comparison between Pure LGCP, Pure RL, BC, and HBRL on the NYC-taxi mobility instantiation ($N{=}6$, 5 seeds). HBRL reaches a higher reward than Pure RL and BC , and reaches Pure RL reward in 
        fewer episodes, reproducing the first instantiation behaviour on a distinct domain using real demand data.}
    \label{fig:mobility_curve}
\end{figure}
We evaluate with the protocol and baselines of Section~\ref{sec:results}, averaged over 5 independent seeds at $N{=}6$. Figure~\ref{fig:mobility_curve} compares learning across all methods, and the ordering reproduces the wireless case: Pure LGCP is stable but non-adaptive, Pure RL and BC converge to similar reward, and HBRL outperforms both, improving over Pure RL by $6\%$ and reaching its final reward level $25\%$ earlier. The gains are smaller than in the wireless case ($10.8\%$ reward, $38\%$ fewer episodes) but identical in direction, indicating a methodological rather than domain-specific benefit; absolute reward magnitudes are not comparable across the two domains, which differ in demand scale and grid resolution, so we compare relative gains only.

\begin{table}[!t]
\caption{Scaling of the converged HBRL reward with fleet size $N$.}
\label{tab:mobility_scaling}
\begin{center}
\begin{tabular}{lccc}
\textbf{Fleet size} & $N = 2$ & $N = 6$ & $N = 10$ \\
\hline \\[-0.9em]
Converged reward & $578$ & $1221$ & $1421$ \\
Reward per vehicle & $289$ & $204$ & $142$ \\
\end{tabular}
\end{center}
\end{table}
\begin{figure*}[!t]
    \centering
    \includegraphics[width=5in]{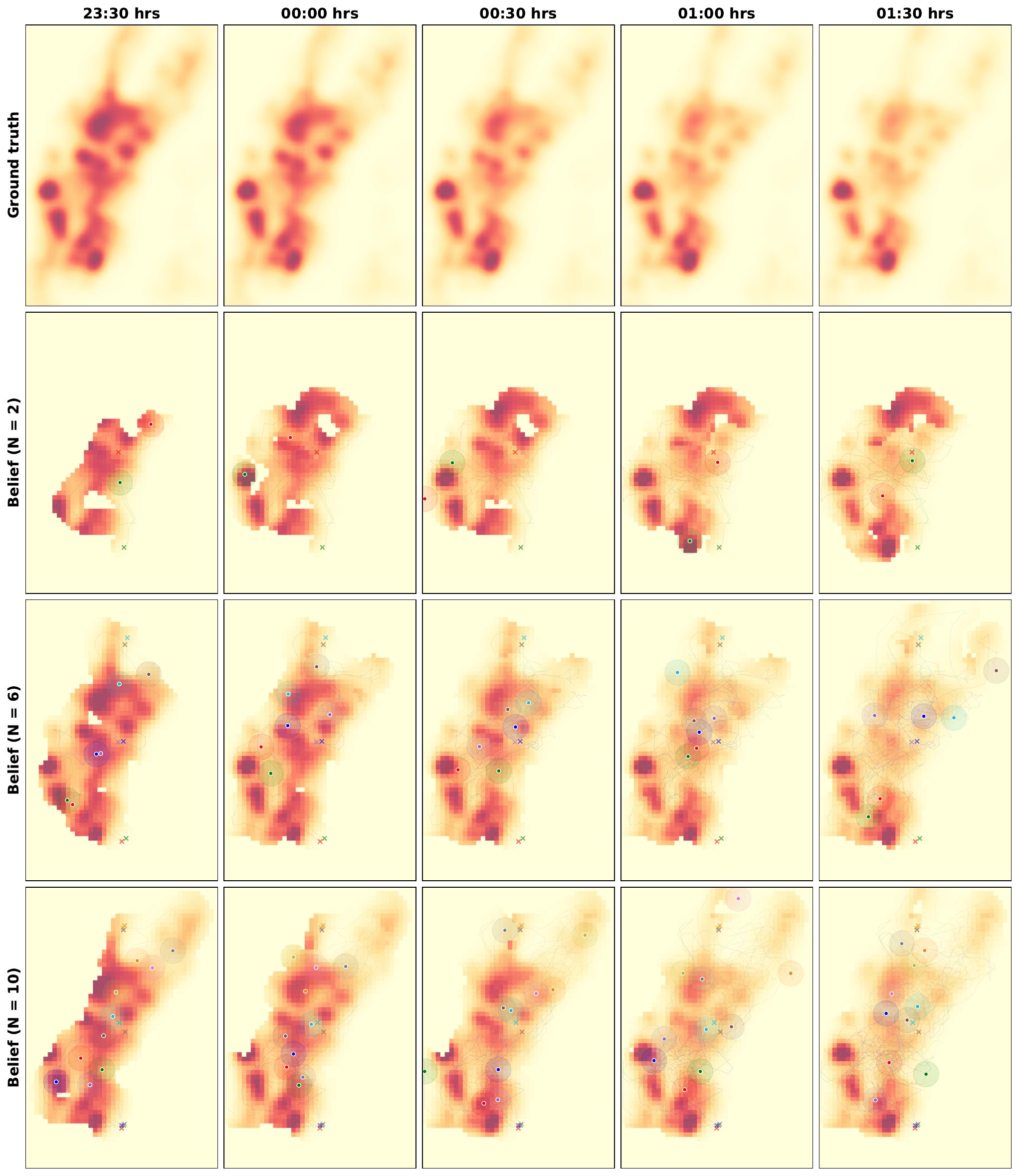}
    \caption{Belief reconstruction versus fleet size over the night episode ($23{:}00$--$02{:}00$); the first snapshot is at $23{:}30$, after the first half-hour of observations. Rows: ground truth and the observed belief for $N = 2, 6, 10$ sensors; columns: successive times. The ground-truth demand (top row) recedes from Midtown toward the downtown hotspots through the night; more sensors yield a fuller reconstruction, sub-linearly.}
    \label{fig:mobility_grid}
\end{figure*}
\subsection{Effect of Fleet size}
The operational area is nearly an order of magnitude larger than the UAV-based wireless provisioning scenario and therefore supports a considerably larger fleet, allowing us to test scalability for larger fleet sizes ($N$ up to $10$, against $N{=}2$--$4$ in Section~\ref{sec:results}). Table~\ref{tab:mobility_scaling} reports how the converged HBRL reward scales with fleet size. Reward increases with $N$ primarily because a small fleet cannot visit, or revisit, the full field within an episode: demand in regions the fleet never reaches is neither observed nor served, and regions visited long ago accumulate uncertainty as the field drifts (Figure~\ref{fig:mobility_grid}). The increase is sub-linear because, as coverage saturates, each added vehicle contributes progressively more redundant coverage, consistent with the scalability trend reported for the wireless case.

Figure~\ref{fig:mobility_grid} shows the belief reconstructed by the fleet over the episode for $N{=}2,6,10$ sensors. Because the demand field evolves through the episode, revisit frequency matters as much as coverage: sparser fleets revisit each region less often, so their belief is staler where the field has since drifted. Down each column the belief fills in as the fleet sweeps, and across columns larger fleets reconstruct more of the field, the qualitative counterpart of the scaling in Table~\ref{tab:mobility_scaling}. At $N{=}2$ the belief is both less complete and more stale, leaving the northern and eastern regions unobserved, while $N{=}6$ and $N{=}10$ revisit frequently enough to track the evolving demand and reconstruct the field almost completely.

\end{document}